\documentclass[man]{apa7}
\usepackage[utf8]{inputenc}
\usepackage[T1]{fontenc}
\usepackage{graphicx}
\usepackage{grffile}
\usepackage{longtable}
\usepackage{wrapfig}
\usepackage[figuresright]{rotating}
\usepackage[normalem]{ulem}
\usepackage{amsmath}
\usepackage{textcomp}
\usepackage{amssymb}
\usepackage{capt-of}
\usepackage{hyperref}
\newcommand{\APABUILD}{}
\makeatletter
\newcommand\efloatredefs{}
\makeatother
\usepackage{floatrow}
\AtBeginDocument{\efloatredefs}
\usepackage[american, english]{babel}
\usepackage{csquotes}
\usepackage[sortcites=true,sorting=nyt,backend=biber,annotation=false,style=apa]{biblatex}
\DeclareLanguageMapping{american}{american-apa}
\usepackage{pgfgantt}
\usepackage{placeins}
\usepackage{float}
\usepackage{listings}
\usepackage{upquote}
\renewcommand{\subparagraph}[1]{\paragraph{\itshape #1}}
\usepackage{booktabs}
\usepackage{threeparttable}
\usepackage{tabularx}
\usepackage{threeparttablex}
\usepackage{siunitx}
\usepackage{makecell}   
\usepackage{multirow}
\usepackage{etoolbox}                
\AtBeginEnvironment{figure}{\let\centering\raggedright}
\AtBeginEnvironment{table}{\let\centering\raggedright}
\AtBeginEnvironment{figure*}{\let\centering\raggedright}
\AtBeginEnvironment{table*}{\let\centering\raggedright}
\usepackage{pdflscape}
\usepackage{caption}
\authorsnames[1, 1, 1, 1, {2,3}, 1, 1]{Dominik Pegler, David Steyrl, Mengfan Zhang, Alexander Karner, Jozsef Arato, Frank Scharnowski,  Filip Melinscak}
\authorsaffiliations{{Department of Cognition, Emotion, and Methods in Psychology, Faculty of Psychology, University of Vienna}, {Vienna Cognitive Science Hub, University of Vienna},{Department of Cognitive Science, Budapest University of Technology and Economics}}
\shorttitle{SpiderNets --- Vision Models Predict Human Fear}
\authornote{Corresponding author: Dominik Pegler. Email: \texttt{dominik.pegler@univie.ac.at}}
\ifdefined\APABUILD
\abstract{Phobias are common and impairing, and exposure therapy, which involves confronting patients with fear-provoking visual stimuli, is the most effective treatment. Scalable computerized exposure therapy requires automated prediction of fear directly from image content to adapt stimulus selection and treatment intensity. Whether such predictions can be made reliably and generalize across individuals and stimuli, however, remains unknown. Here we show that pretrained convolutional and transformer vision models, adapted via transfer learning, accurately predict group-level perceived fear for spider-related images, even when evaluated on new people and new images, achieving a mean absolute error (MAE) below 10 units on the 0--100 fear scale. Visual explanation analyses indicate that predictions are driven by spider-specific regions in the images. Learning-curve analyses show that transformer models are data efficient and approach performance saturation with the available data ($\approx300$ images). Prediction errors increase for very low and very high fear levels and within specific categories of images. These results establish transparent, data-driven fear estimation from images, laying the groundwork for adaptive digital mental health tools.}
\fi
\date{}
\title{SpiderNets: Vision Models Predict Human Fear From Aversive Images}
\hypersetup{
 pdfauthor={},
 pdftitle={SpiderNets: Vision Models Predict Human Fear From Aversive Images},
 pdfkeywords={},
 pdfsubject={},
 pdfcreator={Emacs 31.0.50 (Org mode 9.7.11)}, 
 pdflang={English}}
\begin{document}

\maketitle
\ifdefined\IEEEBUILD
\begin{abstract}
Phobias are common and impairing, and exposure therapy, which involves confronting patients with fear-provoking visual stimuli, is the most effective treatment. Scalable computerized exposure therapy requires automated prediction of fear directly from image content to adapt stimulus selection and treatment intensity. Whether such predictions can be made reliably and generalize across individuals and stimuli, however, remains unknown. Here we show that pretrained convolutional and transformer vision models, adapted via transfer learning, accurately predict group-level perceived fear for spider-related images, even when evaluated on new people and new images, achieving a mean absolute error (MAE) below 10 units on the 0--100 fear scale. Visual explanation analyses indicate that predictions are driven by spider-specific regions in the images. Learning-curve analyses show that transformer models are data efficient and approach performance saturation with the available data ($\approx300$ images). Prediction errors increase for very low and very high fear levels and within specific categories of images. These results establish transparent, data-driven fear estimation from images, laying the groundwork for adaptive digital mental health tools.
\end{abstract}
\fi
\section{Introduction}
\label{sec:org1330eef}

The rapid digitization of healthcare, particularly in mental health, offers new ways to deliver and monitor interventions. In phobias, a prevalent class of anxiety disorders, exposure therapy remains the gold standard \autocites{choyTreatmentSpecificPhobia2007}[][]{craskeMaximizingExposureTherapy2014}[][]{daveyPsychopathologyTreatmentSpecific2007}[][]{healeyExperimentalTestRole2017}. In practice, exposure is conducted along an individualized fear hierarchy: patients are systematically confronted with increasingly challenging stimuli while therapists modulate intensity using reported subjective distress and observed approach or avoidance to optimize therapeutic progress \autocite{craskeMaximizingExposureTherapy2014}. This relies on skilled clinicians, close in-session monitoring, and often resource-intensive in-vivo setups, which limits access and scalability \autocite{choyTreatmentSpecificPhobia2007}. Digital and computerized variants (e.g., image- or virtual reality [VR]-based exposure) promise broader reach and more dynamic, standardized progression, and recent work shows that automated virtual reality exposure therapy (VRET) yields durable benefits under real-world conditions \autocite{miloffAutomatedVirtualReality2019,lindnerGamifiedAutomatedVirtual2020}. Realizing this promise requires reliable, scalable ways to estimate how intense a given stimulus will be at a particular step of the hierarchy --- ideally fast enough to adapt exposures in real time. Computer vision provides tools to estimate fear elicited by visual content. An important step toward adaptive digital exposure therapy is training vision models to predict image-level average fear across individuals \autocite{huysComputationalPsychiatryBridge2016}.

While computer vision research has established a strong foundation for general emotion and sentiment analysis from images \autocite{sunDiscoveringAffectiveRegions2016,youRobustImageSentiment2015,kragelEmotionSchemasAre2019,kimBuildingEmotionalMachines2018,sadeghiDirectPerceptionAffective2024,conwellPerceptualPrimacyFeeling2025,rediesGlobalImageProperties2020}, an important next step is to extend this work to fear-specific stimuli and to integrate explainable artificial intelligence (XAI; \cite{barredoarrietaExplainableArtificialIntelligence2020}). Building on studies that highlight affective regions \autocite{sunDiscoveringAffectiveRegions2016}, there remains an opportunity for quantitative validation of explanations against task-relevant annotations, as saliency can be misleading without rigorous checks \autocite{adebayoSanityChecksSaliency2020}.  Recent work further indicates that improving performance on large-scale vision benchmarks does not by itself ensure that models rely on human-like or biologically plausible visual strategies \autocite{linsleyBetterArtificialIntelligence2025}, making it essential to verify that model attributions are anchored in clinically relevant visual cues rather than arbitrary shortcuts if such systems are to be trusted and adopted in clinical practice.

\begin{figure}[htbp]
  \centering \includegraphics[width=0.8\linewidth]{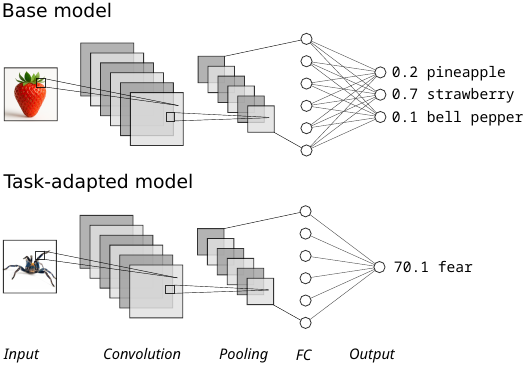}
  \caption{\label{fig:transfer-learning} Transfer Learning from a Pretrained Base Model to a Task-Adapted Model}
\par\footnotesize\textit{Note}. The upper network depicts a pretrained vision model originally trained for multi-class image classification. The diagram uses CNN-style components (convolution, pooling) for illustration; transformers replace these with patch embedding and transformer blocks. "FC" denotes the fully connected layer. The lower network shows a task-adapted model derived from the base model, fine-tuned for a single-output regression task predicting fear ratings. This schematic is conceptual and does not depict the full architectural detail. \end{figure}

We address this gap by adapting pretrained vision models to test how architecture and pretraining shape fear prediction from images, subsequently applying XAI methods. We consider the two dominant architecture families in computer vision, convolutional neural networks (CNNs; \cite{lecunDeepLearning2015}) and vision transformers (ViT; \cite{dosovitskiyImageWorth16x162021}). Within each family, we include one canonical architectural baseline and one more recent representative, spanning both supervised and self-supervised pretraining: ResNet50 \autocite{heDeepResidualLearning2015} (baseline CNN) and ConvNeXtV2 Tiny \autocite{wooConvNeXtV2Codesigning2023} (advanced CNN), and DeiT-S/16 \autocite{touvronTrainingDataefficientImage2021} (baseline transformer) and DINOv2 ViT-S/14 \autocite{oquabDINOv2LearningRobust2024} (advanced transformer, self-supervised). For brevity, we refer to these as ResNet, ConvNeXtV2, DeiT, and DINOv2. The four base models are then fine-tuned on a specialized dataset of spider images to predict human fear ratings \autocite{karnerSpiDaDatasetSelfreport2024}. This approach is known as transfer learning \autocite{yosinskiHowTransferableAre2014} and adapts models trained on large, general datasets \autocite{dengImageNetLargescaleHierarchical2009,oquabDINOv2LearningRobust2024} to new, specific tasks (Figure \ref{fig:transfer-learning}). To mirror real-world use, we evaluate generalization to unseen individuals and unseen images. To support clinical interpretability, we use XAI with quantitative validation to assess whether predictions are grounded in spider-related cues plausibly linked to fear. We generate gradient-weighted class activation mapping (Grad-CAM; \cite{selvarajuGradCAMVisualExplanations2020}) heatmaps and quantify their concentration within spider regions, and we use feature visualization for ResNet \autocite{olahFeatureVisualization2017} to visualize patterns associated with high predicted fear.

Our work makes three contributions: First, we show that pretrained vision models can effectively predict image-level average fear ratings from phobia-specific, spider-related visual stimuli. Second, our explainability analyses confirm that these predictions are grounded in visual cues plausibly related to fear. Third, beyond aggregate accuracy, we identify which image properties are associated with larger prediction errors using an interpretable regression. The combined results provide evidence to improve transparency, guide dataset design, and meet the safety standards for clinical adoption, thereby laying the groundwork for adaptive digital mental health tools.
\section{Methods}
\label{methods}
\subsection{Dataset}
\label{data}
The dataset used in our study was originally collected by \textcite{karnerSpiDaDatasetSelfreport2024} and consists of fear ratings for 313 diverse spider-related images. Ratings were provided on a 0--100 fear scale (0 = no fear, 100 = extreme fear). A total of 152 spider-fearful adults each completed 95 trials, with each trial involving the rating of a single image. The images range from depictions with one or more spiders to scenes with no spiders but only cobwebs, in various contexts including nature, human contact, human environments, and artificial and drawn spiders.

After outlier exclusion (see Supplementary Methods, Appendix \ref{sec:data-details}, for details), the dataset contained 13,532 ratings for 313 images from 148 raters (\(M=\) 43.23 ratings per image, \(SD=\) 12.36, range: 23–87). The overall mean fear rating across images was 50.30 (\(SD=\) 18.98, range: 1.82–83.15). The distributions are illustrated in Figure \ref{fig:data-distributions-all} (Supplementary Results in Appendix \ref{sec:data-distributions}).
\subsection{Base Models and Training}
\label{models-and-training}
The computer vision models used as our \emph{base models} (pretrained networks used as starting points for transfer learning) were ResNet50 \autocite{heDeepResidualLearning2015}, ConvNeXtV2 Tiny \autocite{wooConvNeXtV2Codesigning2023}, DeiT-S/16  \autocites{dosovitskiyImageWorth16x162021}[][]{touvronTrainingDataefficientImage2021}, and DINOv2 ViT-S/14 \autocite{oquabDINOv2LearningRobust2024}. Throughout this paper, we refer to these base models only by their shorter names, ResNet, ConvNeXtV2, DeiT, and DINOv2.

ResNet and ConvNeXtV2 are convolutional neural networks (CNNs; \cite{lecunDeepLearning2015}) pretrained in a supervised way on ImageNet-1K \autocite{dengImageNetLargescaleHierarchical2009}. DeiT and DINOv2 are transformer-based (Vision Transformer [ViT]; \cite{dosovitskiyImageWorth16x162021}); DeiT is supervised \autocite{touvronTrainingDataefficientImage2021} and pretrained on ImageNet-1K, whereas DINOv2 is self-supervised and pretrained on the LVD-142M dataset \autocite{oquabDINOv2LearningRobust2024}. Each pretrained base model is a classifier. We adapted it to single-output regression by replacing the classification head with a small regression head and then applied transfer learning in two stages: partial fine-tuning (head-only) followed by full fine-tuning (all layers). This yielded eight model configurations in total. We refer to each combination of base model and fine-tuning stage (partial or full fine-tuning) as a \emph{model configuration}. Each configuration was trained multiple times within cross-validation (CV), producing multiple trained instances. Images were resized to a common resolution; standard augmentations were used during training; evaluation used resizing only and no augmentations. Implementations used timm and PyTorch \autocites{rw2019timm}[][]{paszkePyTorchImperativeStyle2019}. Full methodological details are provided in Appendix \ref{sec:sup-methods} (Supplementary Methods).
\subsection{Model Evaluation}
\label{model-evaluation}
\subsubsection{Dual-Level Data Partitioning (Participants and Images)}
\label{dual-level-data-partitioning-participants-and-images}
To prevent overlap of the training and held-out test sets at both the participant and image levels, we applied a two-step partitioning scheme. First, we averaged fear ratings for each image separately within two non-overlapping participant groups (A and B). Second, during the 5-fold CV procedure described in the next subsection, images assigned to the training folds were represented only by the group-A means, whereas images in the held-out test fold were represented only by the group-B means. This design ensured that no participant and no image contributed to both training and test data within a fold iteration, preventing overlap and eliminating data leakage.

We randomly split the 148 participants once into two disjoint groups (\(N_A=\) 74 and \(N_B=\) 74) using a fixed random seed and kept this split fixed across all analyses. For each image we computed two targets: group-A mean and group-B mean. In every fold and repetition, models were trained/validated on group-A means for training images and evaluated on group-B means for held-out images. Distributions of group-A and group-B image means are shown in Figure \ref{fig:data-distributions-ab} (Supplementary Results in Appendix \ref{sec:data-distributions}).
\subsubsection{Cross-Validation and Hyperparameter Search}
\label{cross-validation-and-hyperparameter-search}
\begin{figure}[htbp]
  \centering \includegraphics[width=0.8\linewidth]{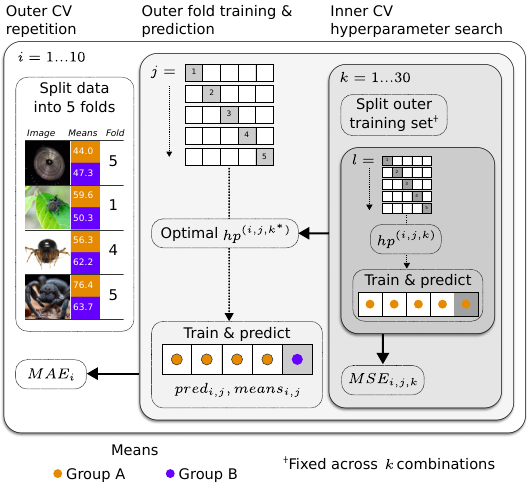}
  \caption{\label{fig:cv} Nested Cross-Validation with Random Hyperparameter Search}
\par\footnotesize\textit{Note}. Indices: $i$ (outer repetitions), $j$ (outer folds), $k$ (30 hyperparameter candidates), and $l$ (inner folds). For each outer repetition $i$, a 5-fold outer split is created. For each outer fold ($j$), hyperparameters (hp) are selected via inner cross-validation over $k$ and $l$. For selection, 5-fold inner CV is run for each $k$, yielding $pred_{i,j,k,l}$. The mean squared error for each hyperparameter candidate is computed as $MSE_{i,j,k} = \text{MSE}(\text{concat}_{l=1}^5 pred_{i,j,k,l}, \text{concat}_{l=1}^5 means_{i,j,k,l})$. The optimal candidate, $k^*$, is chosen as the one with the lowest MSE, then the model is retrained on the full outer training data using $hp^{(i,j,k^*)}$. Final evaluation per repetition is given by $MAE_{i} = \text{MAE}(\text{concat}_{j=1}^5 pred_{i,j}, \text{concat}_{j=1}^5 means_{i,j})$.\end{figure}

We used nested cross-validation with random hyperparameter search to select training settings and estimate out-of-sample performance \autocite{hastie01statisticallearning}. Figure \ref{fig:cv} illustrates the flow of outer and inner folds. Each outer repetition used five folds. Within each outer fold we evaluated 30 hyperparameter candidates by five-fold inner CV. The best inner-CV candidate was then retrained on the full outer training split, and performance was measured on the outer held-out fold.

Models were optimized with AdamW \autocites{loshchilovDecoupledWeightDecay2019}[][]{kingmaAdamMethodStochastic2017} using mean squared error (MSE) as the training loss. At the end of every training epoch we saved a model checkpoint, which is a snapshot of the model parameters at that epoch, and selected the checkpoint with the lowest validation MSE.

Full methodological details are provided in Supplementary Methods  (Appendix \ref{sec:sup-methods}).
\subsubsection{Computation of Final Performance Metrics}
\label{computation-of-final-performance-metrics}
Our primary performance metric is mean absolute error (MAE). We also report root mean squared error (RMSE) and variance explained (\(R^2\)). Evaluation was performed on image-level targets from the held-out rater group (Group B). Predicted scores were clipped to the valid range (0--100) before computing all metrics.

We report two estimators. The single-model estimator computes out-of-fold predictions within each outer repetition and averages the metrics across repetitions. The ensemble estimator averages, per image, out-of-fold predictions across repetitions and then computes the metrics on these averaged predictions to reduce variance \autocite{breimanBaggingPredictors1996,hansenNeuralNetworkEnsembles1990} and stabilize estimates across random data splits and hyperparameter draws. Uncertainty is communicated as 95\% confidence intervals obtained via a hierarchical bootstrap that resamples outer repetitions and images, and mirrors the way the estimators are computed.

To provide a performance ceiling and verify the sufficiency of raters for stable group-level means, we estimated the upper bound on explainable variance by assessing inter-rater agreement using \(\text{ICC(2,k)}\) \autocite{revellePsychProceduresPsychological2024}. This approach quantifies the proportion of between-image variance reliably captured by the group mean and supported our use of a single fixed participant split. Full methodological details are provided in Supplementary Methods (Appendix \ref{sec:sup-metrics}).
\subsection{Learning Curve Analysis}
\label{learning-curve-analysis}
A learning curve analysis quantified how dataset size (number of images available for CV) affects generalization performance on held-out test folds, measured by MAE. Unlike the primary analysis, the learning-curve experiments used partial fine-tuning and single-model estimates for computational feasibility. We ran 15 outer repetitions per dataset size. The analysis consisted of seven different dataset sizes: 50, 75, 100, 150, 200, 250, and 313 (complete dataset) images. For the complete dataset, results were re-used from the primary analysis, and for the other six sizes, each of the 15 outer repetitions used a different random subsample in addition to a different CV split to account for subsampling variability.

We describe learning curves in two simple ways: how fast performance improves and where it levels off. To characterize the relationship between dataset size \(x\) and performance \(y\), we fitted exponential decay models to MAE (\(y(x) = a \cdot exp(-b \cdot x) + c\)) via weighted nonlinear least squares. To make the curves interpretable, we report two milestones: \(x_{50}\) (images needed to achieve half the total improvement) and \(x_{95}\) (images needed for 95\% improvement). Additionally, the asymptotic plateau \(y_{\infty}\) indicates the long-run performance as the dataset grows. Full fitting procedures, parameter definitions, and analyses for RMSE and \(R^2\) appear in Supplementary Methods (Appendix \ref{sec:sup-lc}).
\subsection{Explanations}
\label{explanations}
To make the model's predictions transparent and interpretable, we employed two explainable artificial intelligence (XAI; \cite{barredoarrietaExplainableArtificialIntelligence2020}) approaches: attributions and feature visualization (FV). 
\subsubsection{Attributions}
\label{attributions}
To assess whether model predictions were driven by spider-related content, we examined where models focused when making fear predictions and tested whether attributions concentrated on spider regions. We used gradient-weighted class activation mapping (Grad-CAM; \cite{selvarajuGradCAMVisualExplanations2020}) implemented via the pytorch-grad-cam library \autocite{jacobgilpytorchcam}. Conceptually, Grad-CAM produces a coarse heatmap by weighting the model's final feature representations (just before the regression head) according to how much each location increases the predicted fear, so that higher values indicate locations with greater positive influence on the prediction.

We applied Grad-CAM to all four task-adapted models. For each base model, we used the configuration that achieved the lowest single-model MAE in the primary analysis and generated Grad-CAM heatmaps on held-out test images only, reusing the same outer CV folds and checkpoints as in the performance evaluation (see \hyperref[model-evaluation]{Model Evaluation}). One heatmap was obtained per image and outer repetition, and these were averaged to yield a single heatmap per image and model configuration.

To quantify whether attributions were aligned with spiders, a trained researcher manually annotated spider regions in all images that contained spiders, resulting in masks for 281 of the 313 images. For each annotated image, we summarized Grad-CAM activation inside and outside the spider mask and compared these values across images. We then tested whether mean activation inside the mask exceeded activation outside the mask and reported effect sizes as Cohen's \(d\) for paired samples. Full details are provided in Supplementary Methods (Appendix \ref{sec:attributions-details}).
\subsubsection{Feature Visualization}
\label{feature-visualization}
We performed FV \autocite{olahFeatureVisualization2017} using torch-lucent \autocite{kiatTorchlucent2020} to synthesize input images that maximize the final scalar output (fear) of the task-adapted ResNet model trained on the full dataset with the best-performing training strategy (lowest MAE). We focused on ResNet because, under identical priors and optimization settings, ConvNeXtV2 yielded high-frequency, iridescent patterns with low human interpretability. Adapting FV to ConvNeXtV2 would require alternative, stronger generative priors and is beyond scope. Transformer models were not included due to non-comparable token-level feature spaces. We generated 500 synthetic images for ResNet and 500 ConvNeXtV2 attempts; we show nine ResNet representatives (selected by highest final activation) in Figure 
\ref{fig:xai}b. Full methodological details are provided in Supplementary Methods (Appendix \ref{sec:sup-fv}).
\subsection{Error Analysis}
\label{error-analysis}
We used a linear mixed-effects analysis to relate log-transformed absolute prediction error to four types of predictor: image categories \autocite{karnerVisualAttributesSpiders2024} including spider type and visual attributes (to assess whether certain visual features systematically affect errors), fear ratings (linear and quadratic terms; to test the possibility that errors relate to fear levels and are higher at the extremes), representational distance to the training set (to quantify how typical each image is in the model's learned feature space), and target shift (the absolute difference between Group-B and Group-A image means). Categorical predictors were effect-coded \autocite{singmannIntroductionMixedModels2019} and continuous predictors were z-standardized. We included random intercepts for images to account for the repeated-measures structure in which each image contributes multiple held-out predictions (one per combination of base model and outer repetition). For each base model, we selected the predictions that came from the fine-tuning strategy that demonstrated the best predictive performance in the primary analysis. This yielded \(N=\) 12,520 single held-out predictions (313 images \({\times}\) 10 repetitions \({\times}\) 4 base models). Full details are provided in Supplementary Methods (Appendix \ref{sec:error-details}).
\section{Results}
\label{sec:orgc68773e}
\subsection{Predictive Performance}
\label{cnn-model-performance}
\begin{figure*}[t]
  \centering \includegraphics[width=\textwidth]{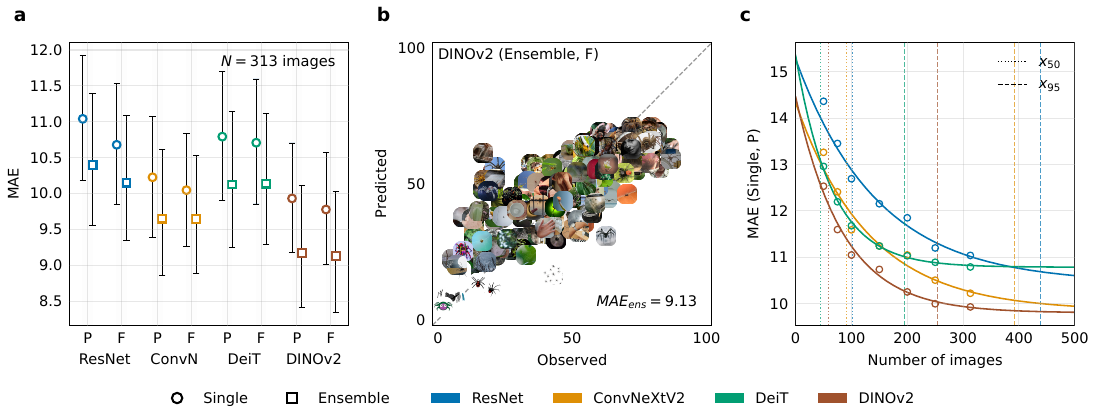}
  \caption{\label{fig:perf-overview} Predictive Performance Overview}
\par\footnotesize\textit{Note}. Panel {\bfseries{a}}: Mean absolute error (MAE) with 95\% bootstrapped confidence intervals (CIs) by base model and prediction mode (single vs. ensemble), showing partial (P) and full (F) fine-tuning. Panel {\bfseries{b}}: observed fear vs. ensemble predictions for the best performing model (DINOv2, fully fine-tuned). The identity line (dashed) represents ideal prediction. For numeric details see Table~\ref{tab:predictions}. Panel {\bfseries{c}}: MAE learning curves for single-model, partial fine-tuning estimates with $x_{50}$ (dotted) and $x_{95}$ (dashed), circles denote empirical points. 
\end{figure*}

As shown in Figure \ref{fig:perf-overview}a, task-adapted single models achieved mean absolute errors (MAEs) ranging from 9.78 to 11.04 (0--100 scale). Among the four base models, DINOv2 and ConvNeXtV2 yielded the lowest MAE, with the best single-model MAE of 9.78 for the fully fine-tuned DINOv2-based model.

We estimated the upper bound on explainable variance via intraclass correlation coefficient (ICC) on a subsample of 74 participants. As shown in Figure \ref{fig:icc}, ICC values stabilized at a high level as rater count increased, placing an upper theoretical bound on \(R^2\) of approximately 0.97. Our best single-model and ensemble \(R^2\) values (0.61 and 0.66) lie below this bound. The stabilization indicates that the number of ratings per image is sufficient and supports using a single fixed participant split for evaluation.

Ensembling the models consistently reduced error across all base-model configurations. For instance, the ensemble of fully fine-tuned DINOv2-based models achieved the lowest MAE of 9.13 --- an absolute decrease of 0.65 in error over the best single model (9.78), corresponding to a 6.6\% relative improvement. Figure \ref{fig:perf-overview}b illustrates observed fear versus ensemble predictions for the best-performing configuration (DINOv2 with full fine-tuning). 

Transformers and CNNs differed in their learning curves (Figure \ref{fig:perf-overview}c). Transformers (DeiT, DINOv2) improved quickly and were effectively saturated when using the complete dataset (313 images), with 50\% of the total improvement (\(x_{50}\)) reached after 45–59 images and \(x_{95}\) after 194–255. CNNs (ResNet, ConvNeXtV2) required about 91–102 images to reach \(x_{50}\) and about 392–439 to approach \(x_{95}\), indicating modest headroom at the present dataset size. The fitted MAE plateaus were lowest for DINOv2 (9.80) and ConvNeXtV2 (9.85).

Training time for the full dataset per outer repetition (5-fold outer CV with inner-CV hyperparameter search) on our NVIDIA RTX 4080 GPU ranged from 4.7–7.8 h for partial fine-tuning and from 14.7–21.3 h for full fine-tuning (fastest: ResNet, slowest: DeiT). Full fine-tuning times include the partial stage.

Detailed numerical results, training-time summaries (Table \ref{tab:predictions}; Figures \ref{fig:performance-bars}, \ref{fig:predictions}) and the full learning-curve analyses (Figure \ref{fig:learning-curves}; Tables \ref{tab:learning-curve-convergence}, \ref{tab:empirical-learning-curves}, \ref{tab:learning-curve-params}) appear in Supplementary Results (Appendix \ref{sec:predictive-performance}).
\subsection{Model Explanations}
\label{sec:orgb26be55}

\begin{figure*}[t]
  \centering \includegraphics[width=\textwidth]{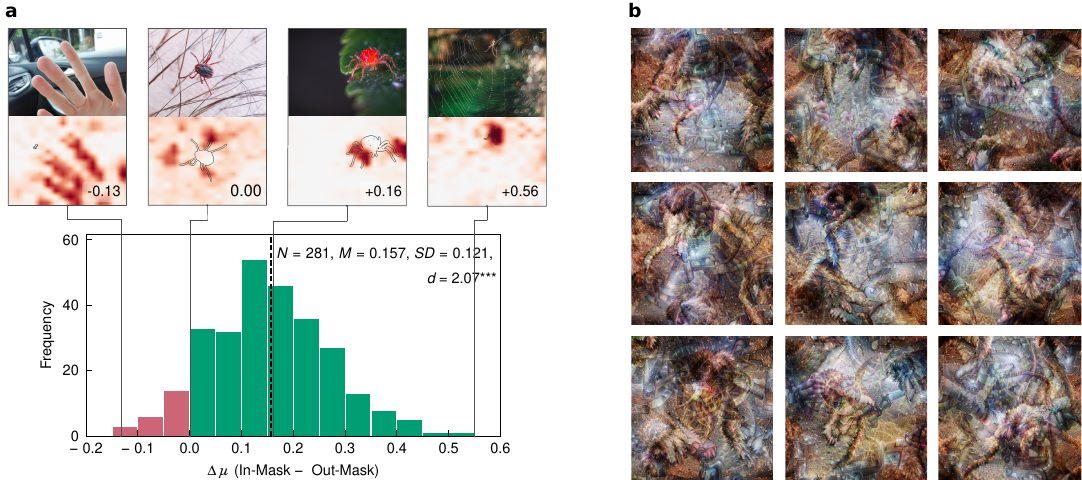}
  \caption{\label{fig:xai} Explainability Results: Grad-CAM and Feature Visualizations}
\par\footnotesize\textit{Note}. Panel {\bfseries{a}}: Four example Grad-CAM heatmaps for held-out spider images and their position in the distribution of $\Delta\mu$ ($\mu_{in} - \mu_{out}$; mean activation inside minus outside the annotated spider mask) for the best performing model configuration (DINOv2, fully fine-tuned). Positive $\Delta\mu$ indicates greater focus on spider regions; negative indicates focus elsewhere. Green areas mark the proportion of images with higher Grad-CAM activation inside spider masks than outside; red marks the opposite.  Asterisks denote paired one-sided $t$-tests (inside $>$ outside): * $p<0.05$; ** $p<0.01$; *** $p<0.001$; $d$ is Cohen's $d$. Panel {\bfseries{b}}: Nine synthetic inputs that maximally activate the task-adapted ResNet's fear output (selected by highest activation). A total of 500 such images were generated, 83.8\% of which were classified as "tarantula" by an unadapted ResNet (Figure \ref{fig:fv-classes-resnet}).
\end{figure*}

Gradient-weighted class activation mapping (Grad-CAM) analyses revealed that all four base models predominantly rely on spider regions when predicting fear. Figure \ref{fig:xai}a shows the distribution of activation differences and example heatmaps for the best-performing model (DINOv2), illustrating concentration of attention within spider areas. Quantitative analysis confirmed this pattern across all models: using raw Grad-CAM scores without per-image rescaling, the paired differences \(\Delta\mu = \mu_{in}-\mu_{out}\) were positive on average, with effect sizes ranging from large (DeiT) to very large (DINOv2) (see Figure \ref{fig:gradcam-difference-distribution} for per-model details). Distributions and example heatmaps for all base models can be found in Appendix \ref{sec:gc}. The full set of these heatmaps is available at \url{https://osf.io/8b5a2}.

Feature visualizations (FVs) for the ResNet-based model show coherent spider-like structure (Figure \ref{fig:xai}b); 83.8\% (95\% CI 80.3\%–86.8\%) of these generated images were classified as ``spider'' (top-1) by an unadapted ResNet-ImageNet classifier (\(N=\) 500); summary statistics are provided in Table \ref{tab:fv-summary}. In contrast, ConvNeXtV2 FV attempts were not readily interpretable and reached 13.0\% (95\% CI 10.3\%–16.2\%) top-1 ``spider'' classification rate (\(N=\) 500).
\subsection{Error Analysis Results}
\label{sec:org321b273}

\begin{table*}[htbp]
\centering
\caption{Omnibus Wald Tests for Categorical Predictors}
\label{tab:anova-omnibus}
\begin{threeparttable}
\fontsize{11}{13}\selectfont
\begin{tabular}{lllS[table-format=1.3]S[table-format=1.3]S[table-format=2.3]l}
\toprule
\addlinespace[0.3em]
Factor & Categories & $k$ & {$df$} & {Wald $\chi^2$} & $p$ \\
\midrule
Model & ResNet, ConvNeXtV2, DINOv2, DeiT & 4 & 3 & 35.061 & <.001 \\
Environment & nature, civilization, human contact, not evaluable & 4 & 3 & 10.863 & 0.012 \\
Cobweb & absent, present & 2 & 1 & 6.158 & 0.013 \\
Distance & close, distant & 2 & 1 & 5.580 & 0.018 \\
Texture & smooth, hairy & 2 & 1 & 3.353 & 0.067 \\
Prominent legs & no, yes & 2 & 1 & 2.546 & 0.111 \\
Perspective & side/front, top/bottom & 2 & 1 & 0.624 & 0.430 \\
Spider type & no spider, real spider, artificial spider & 3 & 2 & 1.227 & 0.542 \\
Eating prey & not eating, eating & 2 & 1 & 0.347 & 0.556 \\
Number of spiders & one, two+ & 2 & 1 & 0.253 & 0.615 \\
Color & color, bw & 2 & 1 & 0.151 & 0.697 \\
Eyes & non-visible, visible & 2 & 1 & 0.092 & 0.761 \\
Size & small, middle, large & 3 & 2 & 0.317 & 0.854 \\
\bottomrule
\end{tabular}

\begin{tablenotes}
\item \textit{Note.} Linear mixed-effects regression on log absolute error with random intercepts for images. Categorical factors are effect-coded (sum-to-zero). The Wald $\chi^2$ statistic tests whether all level coefficients for a factor jointly equal zero; $k$ = number of levels, $df$ = degrees of freedom. $p$-values are unadjusted.
\end{tablenotes}
\end{threeparttable}
\end{table*}

\begin{figure}[htbp]
  \centering \includegraphics[width=\textwidth]{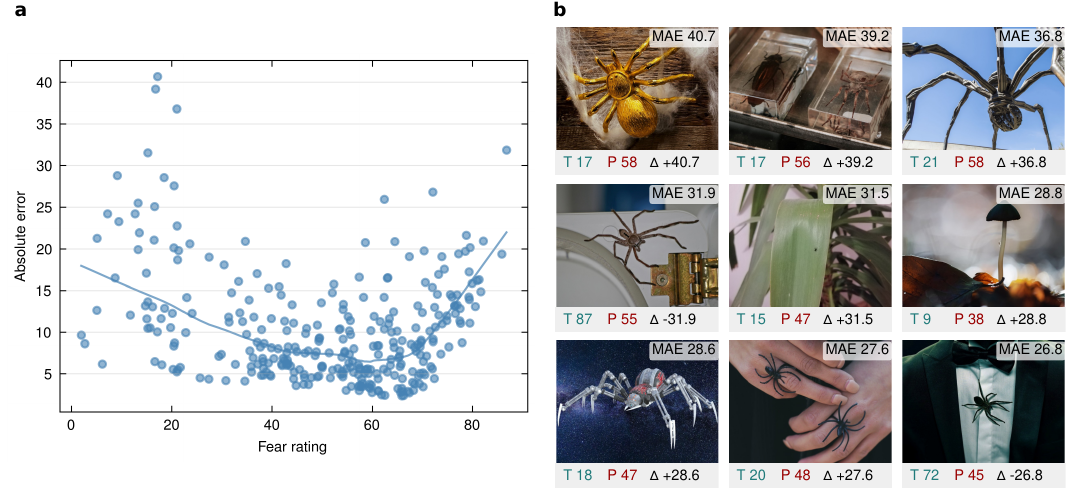}
  \caption{\label{fig:err} Absolute Error vs. Fear, and Highest-Error Images}
  {\footnotesize\textit{Note}. Panel {\bfseries a}: Absolute error vs. held-out fear. Points show image-level mean prediction errors; lines are locally weighted scatterplot smoothing (LOWESS) fits. Panel {\bfseries b}: Nine highest-error images across all configurations with mean absolute/signed errors (MAE/$\Delta$), true fear (T), and mean predicted fear (P).}
\end{figure}

We fitted a single linear mixed-effects model that explained considerable variance in log absolute prediction errors (\(R^2=\) 0.42). Table \ref{tab:anova-omnibus} summarizes omnibus Wald tests for the categorical predictors; exact coefficients of all predictors appear in Table \ref{tab:mixed-effects-full}. The strongest effect was a positive quadratic fear term, yielding a convex (U-shaped) pattern (Figure \ref{fig:err}a) with larger errors toward very low and very high fear levels. Environment showed a reliable omnibus effect, driven by higher errors for images set in civilization contexts and lower errors for ``not evaluable'' backgrounds relative to the grand mean. Cobweb presence also had a significant omnibus effect and was associated with lower errors when cobwebs were present. Disagreements between training and held-out rater groups increased prediction errors. Distance showed an omnibus effect, with distant viewpoints tending toward higher errors than close ones when spiders were present, although level-specific contrasts were small and only borderline after correcting for multiple comparisons.

Figure \ref{fig:err}b shows the nine images (\(\approx\)3\% of the dataset) with the highest mean absolute errors when averaged across all model configurations; together they account for \(\approx\)9\% of the total absolute error.
\section{Discussion}
\label{sec:org3a56e30}

This study investigated whether pretrained vision models can predict how strongly spider-related images evoke fear in humans. Across convolutional neural network (CNN; \cite{lecunDeepLearning2015}) and transformer-based (\cites{dosovitskiyImageWorth16x162021}[][]{touvronTrainingDataefficientImage2021}) architectures adapted via transfer learning, single models achieved a mean absolute error of 9.78 on the 0--100 fear scale (ensemble 9.13) when applied to new images and new raters. Transformers tended to outperform CNNs and, according to learning-curve analyses, were effectively saturated at the current dataset size, whereas CNNs retained modest headroom. Ensembling and full fine-tuning yielded small but consistent gains. Explainability analyses further indicated that the models' fear predictions were grounded in spider-specific regions, suggesting that they relied on clinically relevant cues rather than opaque background patterns, supporting their trustworthiness and safety for potential therapeutic use (Figure \ref{fig:xai}). Together, these results indicate that pretrained vision models can provide accurate and interpretable estimates of image-evoked fear. Below, we relate these findings to prior work on affective computer vision, consider their implications for exposure-based treatments, and outline key limitations and directions for future research.
\subsection{Results in Context of Affective Computer Vision}
\label{sec:org207d323}

Across architectures, our models demonstrated robust predictive power for image-evoked fear, aligning these findings with the broader literature on computational models of affect derived from visual input. Prior work on models of image-based affect estimation has followed three main approaches, which differ in how visual information is represented and mapped to affective judgments: (1) handcrafted global properties (e.g., color, texture), which summarize low-level image statistics, predict continuous valence ratings with a modest share of variance (adjusted \(R^2 \approx 0.20\)) \autocite{rediesGlobalImageProperties2020}; (2) engineered perceptual descriptors (e.g., spatial-frequency content, summaries of early CNN layers), which approximate early stages of visual processing, yield valence correlations of up to \(r \approx 0.47\) \autocite{sadeghiDirectPerceptionAffective2024}; and (3) deep learning approaches --- including end-to-end predictors, deep-feature pipelines, and region-based fusion models that isolate ``affective regions'' \autocite{sunDiscoveringAffectiveRegions2016} --- achieve substantially higher performance across affect datasets compared to the aforementioned approaches. For instance, regression on the International Affective Picture System (IAPS) yielded \(r \approx 0.88\) for valence and \(r \approx 0.85\) for arousal \autocites{kragelEmotionSchemasAre2019}[][]{kimBuildingEmotionalMachines2018}, while binary sentiment classification for images sourced from Twitter reached accuracies up to 88.9\% \autocite{sunDiscoveringAffectiveRegions2016}. More recent reliability-aware comparisons demonstrate that large-scale pretrained visual representations (both vision-only and vision–language) explain a considerable share of the variance in group-average affect and aesthetics ratings: reliability-adjusted explained variance reaches approximately 0.73 for state-of-the-art vision-only models and up to 0.87 for vision–language models \autocite{conwellPerceptualPrimacyFeeling2025}.

Our work falls in the third category: pretrained vision models, adapted via transfer learning, accurately predict phobia-specific fear on a 0--100 scale at the group level under strict out-of-sample evaluation that holds out both images and raters. The best single model achieved \(R^2=\) 0.61, with an ensemble reaching \(R^2=\) 0.66. Direct numerical comparison with prior studies is constrained mainly by differences in constructs (fear vs. broad valence/arousal or aesthetics) and rating scales. Conceptually, however, our findings align with the view that high-level visual features learned from large image corpora transfer well to affective judgments at the group level \autocite{kragelEmotionSchemasAre2019,kimBuildingEmotionalMachines2018,conwellPerceptualPrimacyFeeling2025}, and extend these observations to fear elicited by clinically relevant, phobia-specific images.
\subsection{Architecture Choice and Data Efficiency}
\label{sec:org62884a5}

Compared to CNNs, transformer-based models were more data efficient at the present dataset size, with learning curves indicating that DeiT and DINOv2 were already close to their performance asymptotes, whereas ResNet and ConvNeXtV2 continued to show modest headroom (Figure \ref{fig:perf-overview}b). Overall, DINOv2 and ConvNeXtV2 achieved the lowest fitted error plateaus, consistent with advances in both architecture design and large-scale pretraining.

These architectural differences align with the broader move toward transfer learning and foundation-style models in data-constrained domains, where carefully labeled datasets are typically small and expensive to obtain. Transformers such as DeiT and DINOv2 leverage large-scale supervised or self-supervised pretraining \autocites{touvronTrainingDataefficientImage2021}[][]{oquabDINOv2LearningRobust2024} to provide general visual features that can be adapted with comparatively few task-specific examples. Modern CNNs can remain competitive and in some regimes match or surpass transformers when sufficient data are available \autocite{vishniakovConvNetVsTransformer2024}, but in phobia-specific fear prediction with only a few hundred labeled images, our results suggest that moderately sized pretrained transformers offer a particularly favorable balance between accuracy, data efficiency, and computational cost. At the same time, the remaining headroom for ConvNeXtV2 indicates that, if larger, well-curated phobia image sets become available, strong CNNs may still benefit from additional data and could even exceed transformer performance. Large-scale collaborative initiatives \autocite{hebartTHINGSDatabase18542019,hebartTHINGSdataMultimodalCollection2023} and the growing adoption of open science and reproducible research practices \autocite{munafoManifestoReproducibleScience2017} make the future availability of such datasets plausible, facilitating the development of shared, well-curated phobia image resources that support data-intensive models and enable systematic cross-study benchmarking.
\subsection{Interpretability, Error Patterns, and Dataset Implications}
\label{sec:org964bea5}

Accuracy alone is insufficient for clinical uptake \autocite{linsleyBetterArtificialIntelligence2025}. For AI-based tools that might inform exposure therapy, clinicians need to know which visual cues drive predictions, whether these cues are clinically meaningful, and where the models are likely to fail. We therefore combined attribution-based and generative explainability methods with a systematic error analysis.

On the interpretability side, gradient-weighted class activation mapping (Grad-CAM; \cite{selvarajuGradCAMVisualExplanations2020}) showed that, across architectures, attributions were reliably higher inside spider regions, than in unrelated background regions (Figure \ref{fig:xai}; Figure \ref{fig:gradcam-difference-distribution}). Feature visualizations (FVs; \cite{olahFeatureVisualization2017}) for the task-adapted ResNet further produced coherent spider-like structure, with most synthetic images classified as spiders by an unadapted ImageNet model. Within the broader explainable AI literature \autocite{barredoarrietaExplainableArtificialIntelligence2020}, this moves beyond purely qualitative saliency maps by quantitatively validating attributions against task-relevant annotations, and addresses concerns that saliency methods can be misleading when left unchecked \autocite{adebayoSanityChecksSaliency2020}.

The error analysis complements these findings by clarifying where generalization is most fragile. Absolute errors were larger at very low and very high fear, yielding a convex pattern across the scale (Figure \ref{fig:err}). A likely contributor is the stimulus distribution: image-level fear means were approximately normal, with many mid-range stimuli and comparatively few near 0 or 100 (Figures \ref{fig:data-distributions-all} and \ref{fig:data-distributions-ab}). Thin coverage at the tails provides limited information to constrain the models and increases uncertainty in those regions. Beyond fear level, errors followed structured patterns. Images in which spiders were visually prominent and easy to identify tended to yield smaller errors. In contrast, errors were higher for visually complex or ambiguous scenes, especially when spiders appeared in non-natural or unusual contexts (e.g., artificial depictions or jewelry such as a golden spider).

Taken together, the explanation and error analyses delineate where fear predictions can be relied on and where they should be treated with caution. The attribution results show that models typically base their predictions on spider-related content, which is a necessary condition for clinical acceptability. The error patterns show that this is not sufficient: predictions degrade in systematic ways, especially at very low and very high fear levels and in visually complex or unusual scenes. These patterns have direct implications for the design and evaluation of future systems. They point to the need for better coverage of underrepresented regions of the fear distribution, for targeted stress tests in visually demanding contexts, and for architectures that can better handle ambiguity, such as models that combine visual and textual input \autocite{radfordLearningTransferableVisual2021}. 
\subsection{Limitations}
\label{sec:orgab46c11}

We note several limitations. First, the present findings are bounded by the stimulus domain (spider-related images) and by group-level targets (image-level mean fear), which do not capture individual variability; personalization will be necessary for clinical use. Predicting a single person's fear rating for an image leads to higher errors than predicting the average rating for that image across individuals. This occurs because the variability in individual responses to the same stimulus adds noise when using a model trained on group-level means to predict individual ratings (see \hyperref[sec:individual-errors]{Individual- vs. Group-Level Errors} in Supplementary Results). Second, the distribution of image-level fear means was approximately normal rather than uniform across the 0--100 range, with many mid-range stimuli and comparatively few at the tails (Figures \ref{fig:data-distributions-all} and \ref{fig:data-distributions-ab}). This imbalance reduces training data at very low and very high fear and may have contributed to larger errors in those regions, so future datasets should aim for a more uniform coverage. Third, our cross-split ensembling yields optimistic gains relative to same-split ensembling. Observed improvements likely reflect both variance reduction from ensembling \autocites{breimanBaggingPredictors1996}[][]{hansenNeuralNetworkEnsembles1990} and increased training data size across outer repetitions; accordingly, ensemble numbers are best viewed as upper-end estimates.  Finally, our explanations were partially constrained by applying FV to ResNet only, which limits generalizability, and we did not conduct broader independent validation of interpretability results (e.g., agreement with a separate classifier on object identity or structured human ratings).
\subsection{Clinical Relevance and Future Directions}
\label{sec:orge722c00}

In clinical terms, the achieved accuracy can inform core operations of exposure therapy, in which visual stimuli are selected and sequenced along graded hierarchies \autocites{craskeMaximizingExposureTherapy2014}[][]{choyTreatmentSpecificPhobia2007}. On a 0--100 fear/subjective units of distress scale, adjacent steps are typically defined by modest differences, and our single-model errors of about 9.78 (variance explained 0.61) appear sufficient to support the ordinal decisions clinicians make when constructing and advancing hierarchies. These hierarchies emphasize relative ordering and stepwise calibration to current subjective units of distress and observed approach–avoidance \autocite{craskeMaximizingExposureTherapy2014}. Under clinician oversight, model predictions can be used to (i) propose and order candidate images at nearby steps, (ii) flag potential outliers for review, and (iii) adjust progression in real time using conservative decision margins, in line with requirements for automated exposure therapy in real-world settings \autocites{miloffAutomatedVirtualReality2019}[][]{lindnerGamifiedAutomatedVirtual2020}. For clinical deployment, uncertainty-aware safeguards are needed --- such as stochastic variance estimates and out-of-distribution detection --- that allow the system to abstain and defer to clinician judgment in cases of high predictive uncertainty \autocites{galDropoutBayesianApproximation2016}[][]{begoliNeedUncertaintyQuantification2019}.

To enhance clinical utility further, a key next step is to move from group-level predictions toward individualized assistance that incorporates patient-specific information. One route is to fuse image-derived features with non-visual covariates (such as demographics, baseline symptom severity, or questionnaire scores) in a small multimodal prediction head that conditions fear estimates on patient-level information, or to insert lightweight adapter layers that modulate the vision model using covariate embeddings \autocites{raghuRapidLearningFeature2020}[][]{liMultimodalAlignmentFusion2025}. These two approaches keep the pretrained vision model largely fixed, support rapid personalization from modest numbers of patient-specific ratings, and are compatible with established multimodal learning strategies for clinical prediction models \autocites{rajkomarMachineLearningMedicine2019}[][]{celikorsBeautyEyeYour2025}.

The same framework can extend beyond arachnophobia to other specific phobias (for example, other animal phobias, heights, or needles and blood) given suitable domain-specific image corpora, safety review, and rater protocols aligned with clinical practice \autocite{choyTreatmentSpecificPhobia2007}. Moving from static images to dynamic content in virtual reality (VR) would further enable moment-to-moment estimation of fear within immersive sessions. Contemporary video transformers such as VideoMAE, which learn temporal representations from limited labeled data, appear promising in these data-constrained settings \autocite{tongVideoMAEMaskedAutoencoders2022}. At the same time, our error patterns suggest that the stimulus space should be deliberately enriched: targeted data collection strategies (for example, adaptive sampling or active learning) could prioritize stimuli likely to elicit very low or very high fear, thereby improving coverage at the extremes of the fear distribution where current predictions are least reliable \autocite{roadsEnrichingImageNetHuman2020}.
\subsection{Conclusion}
\label{sec:org5d18f7b}

In conclusion, pretrained vision models estimated group-level fear ratings from spider images with substantial accuracy and interpretable visual grounding. Within research on affective computer vision, these results align with evidence that visual features can meaningfully predict affective judgments and extend this work to phobia-specific fear. The findings provide the foundation for clinician-supervised exposure therapy workflows by enabling informed stimulus selection and graded progression. They motivate deployment strategies that explicitly account for prediction uncertainty, support personalization from modest numbers of patient ratings, extend to other phobia domains and to video-based VR, and maintain transparent, quantitatively validated explanations. Together, the findings represent a step toward safety-aware, image-based assistants that can help make exposure therapy more adaptive and scalable.
\section{Data Availability Statement}
\label{data-availability-statement}
The data used in this study are available from previous work. The spider images are from \textcite{zhangSpiDaMRIBehavioralFMRI2025}, the fear ratings for these images are from \textcite{karnerSpiDaDatasetSelfreport2024}, and the categorization taxonomy for the images is from \textcite{karnerVisualAttributesSpiders2024}. The code and materials to reproduce the analysis are available at \url{https://osf.io/8b5a2/}.
\section{CRediT Authorship Contribution Statement}
\label{credit-authorship-contribution-statement}
DP: Conceptualization, Investigation, Methodology, Software, Formal Analysis, Data Curation, Visualization, Writing -- Original Draft, Writing -- Review \& Editing. DS: Methodology, Writing -- Review \& Editing. MZ: Data Curation, Writing -- Review \& Editing. AK: Data Curation, Writing -- Review \& Editing. JA: Formal Analysis, Writing -- Review \& Editing. FS: Resources, Supervision, Writing -- Review \& Editing. FM: Supervision, Conceptualization, Methodology, Writing -- Original Draft, Writing -- Review \& Editing.
\section{Acknowledgments}
\label{acknowledgments}
We would like to thank Jázmin Mikó for significant contributions to the spider labeling, which were instrumental in advancing this project.
\section{Competing Interests}
\label{competing-interests}
The authors declare no competing interests.
\section{Funding}
\label{funding}
This research was funded by the Austrian Research Promotion Agency (FFG), Project Nos. 927913 \& 887474, and the European Union's Horizon Europe research and innovation programme under grant agreement ID 101155881. FM was funded by the Austrian Science Fund (FWF) [10.55776/ESP133].

\FloatBarrier
\section{References}
\label{references}
\printbibliography[heading=none]

\clearpage
\appendix
\FloatBarrier
\clearpage
\section{Supplementary Methods}
\label{sec:org03911c8}
\label{sec:sup-methods}
\subsection{Dataset Details}
\label{sec:orgc3d176b}
\label{sec:data-details}

As described in the main text, the raw dataset from \textcite{karnerSpiDaDatasetSelfreport2024} comprised 313 diverse spider-related images rated on a 0--100 fear scale by 152 spider-fearful adults, each completing 95 trials with one image per trial. Within each participant's session, a subset of images was repeated to assess within-participant reliability. For the present analyses, and following \textcite{karnerVisualAttributesSpiders2024}, we retained only the rating from the first presentation of each image per participant when preparing the dataset for model training and evaluation.

After restricting to first presentations, the dataset contained 13,891 ratings. Each image was rated by varying numbers of participants, with a mean of 44.38 ratings per image (\(SD=\) 12.64, range: 24–88).

We then identified and removed participant outliers using two complementary methods applied to within-image ratings. First, we used a correlation-based method to assess agreement with the other raters. We computed the median fear rating for each image across all raters (group-median ratings). For each participant, we then computed Spearman's rank correlation between that participant's ratings and the group-median ratings across the same images, which captures agreement in the relative ordering of images irrespective of absolute rating levels. Participants with a correlation coefficient below the threshold defined as the first quartile (Q1) minus 1.5 times the interquartile range (IQR) of all correlation coefficients were flagged as outliers, indicating substantial rank-order deviation relative to the ordering implied by the group-median ratings.

Second, we implemented an approach based on the median absolute deviation \autocite{leysDetectingOutliersNot2013}. We calculated the median rating for each image and then determined the absolute deviations of individual ratings from these medians. By computing the median absolute deviation for each participant, we identified those with deviations greater than the threshold defined as the third quartile (Q3) plus 1.5 times the IQR. Participants were excluded if flagged by at least one of the two criteria.

Before outlier exclusion, the dataset thus comprised 13,891 ratings for 313 images. After applying both outlier criteria, we identified and excluded 4 participants who contributed a total of 359 ratings. The remaining dataset, used in all subsequent analyses, contained 13,532 ratings for 313 images from 148 raters. Distributions are illustrated in Supplementary Results (Appendix \ref{sec:data-distributions}).
\subsection{Base Models and Training Details}
\label{sec:org3058c87}
\label{sec:sup-base-models}

ResNet is a widely used convolutional neural network (CNN; \cite{lecunDeepLearning2015}) baseline that has demonstrated strong performance across various applications, including adaptation for regression problems and learning latent variables \autocite{sandersTrainingDeepNetworks2020,sandersVisionlanguageModelsLearn2025}. ConvNeXtV2 represents a modern CNN architecture that has proven effective in specialized tasks such as spider classification \autocite{dengLookOutDangerous2024}. DeiT and DINOv2 use a modern transformer architecture, which allows us to compare classic vision models (CNNs) with increasingly relevant transformer-based models. DeiT serves as our canonical transformer baseline, representing the supervised learning paradigm, similar to our chosen CNN models. In contrast, DINOv2 is the only self-supervised base model in our set. While both DeiT and DINOv2 are built on the same core Vision Transformer (ViT; \cite{wuVisualTransformersTokenbased2020,dosovitskiyImageWorth16x162021}) architecture, DINOv2 differs significantly by its pretraining method --- it used no human labels for feature learning.

All four base models are moderately sized (ResNet50: 25.6M parameters, ConvNeXtV2 Tiny: 28.6M, DeiT-S/16: 22.1M, DINOv2 ViT-S/14: 22.1M), ensuring that they remain manageable with limited computational resources, which is crucial for potential deployment in clinical settings, including on edge devices or systems with constrained processing capabilities. Another important reason for selecting these models was the availability of pretrained weights, specifically the standard ImageNet-1K weights for the supervised models \autocite{dengImageNetLargescaleHierarchical2009}, with the exception of the self-supervised DINOv2, which was trained on the large-scale LVD-142M dataset \autocite{oquabDINOv2LearningRobust2024}.

We replaced each base model's classification head with a regression head comprising three fully connected hidden layers (256 units each) and a final linear output. The feature vector from the base model is first passed through dropout and projected to 256 units. Each subsequent hidden layer applies rectified linear unit (ReLU) activation, batch normalization, dropout, and a 256-unit fully connected transformation. Immediately before the output, ReLU activation, batch normalization, and dropout are applied, followed by a linear unit that produces a single scalar. No activation is applied after the final layer. Clipping was not used during training or validation; losses were computed on raw, unclipped outputs, and clipping to the 0--100 rating range was applied only to the final held-out predictions. We refer to the resulting network as the \emph{task-adapted model}.

For fine-tuning, all 313 spider-related images were processed by an input pipeline that resized each image to 224\texttimes{}224 pixels at input time. For DINOv2, we used the ViT-S/14 reg=4 (four register tokens) variant at 224\texttimes{}224 and interpolated positional embeddings from its original 518\texttimes{}518 pretraining resolution to maintain a uniform input resolution and batch size across models \autocite{oquabDINOv2LearningRobust2024}. Labels were image-level fear ratings from our dataset. To enhance robustness and generalizability, the pipeline applied standard image augmentations during training mode \autocite{shortenSurveyImageData2019}: random horizontal flips, small rotations, translations, shearing, scaling, color jittering, filling, and Gaussian blurring.

Our training involved two stages --- partial and full fine-tuning. In partial fine-tuning, pretrained layers were frozen and only the new head was trained. In full fine-tuning, the previously frozen pretrained layers were unfrozen, and training continued with all layers updated on the partially fine-tuned model. Each base model was evaluated under two configurations: partial and full fine-tuning. This yielded eight model configurations and makes the partially and fully fine-tuned models paired within a given outer cross-validation fold rather than independent runs.

All four base models were sourced from the timm Python library \autocite{rw2019timm} with pretrained weights and implemented in PyTorch \autocite{paszkePyTorchImperativeStyle2019}. Training was conducted on an NVIDIA RTX 4080 GPU with 16 GB VRAM.
\subsection{Nested Cross-Validation and Hyperparameter Search}
\label{sec:org5b07810}
\label{sec:sup-cv}

The study implemented a multi-stage nested CV and hyperparameter search procedure (Figure \ref{fig:cv}) to optimize and evaluate the performance of the models \autocite{hastie01statisticallearning}. This process included four nested loops. The first loop performed 10 repetitions of random splits, each creating five folds, to ensure robust results. The second loop, the outer CV, then used these same splits for training and evaluation across the five outer folds. The third loop iterated over 30 randomly sampled hyperparameter candidates per outer fold (random search). The fourth loop (inner CV) was a 5-fold CV used to select the best candidate by lowest inner-CV loss. All 30 candidates were evaluated on the same set of 5 inner folds to ensure comparability. In each outer repetition, the 5-fold outer CV produced five task-adapted model instances per model configuration, each trained on four folds and evaluated on its held-out test fold. Across the 10 repetitions, this yielded 50 trained instances per configuration. Each image received exactly one held-out prediction per repetition (10 in total).

The batch size was fixed at 16, and the AdamW optimizer \autocites{kingmaAdamMethodStochastic2017}[][]{loshchilovDecoupledWeightDecay2019} was used. Mean squared error (MSE) was used as both the loss function during training and as the metric for evaluating model performance at each epoch. To avoid gradient saturation and bias from bounded output constraints, we left the final unit unconstrained and computed MSE on raw predictions. Clipping to 0–100 was applied only post hoc, and we stored both raw and clipped predictions. At the end of each epoch, evaluation was conducted on an internal validation set, which consisted of 20\% of the training data (group-A means). The checkpoint (a snapshot of the model's weights) corresponding to the lowest MSE on this validation set was selected as the best-performing model instance. We trained to a preset epoch cap without early stopping or learning rate schedulers; predictions use the best-validation checkpoint within those epochs.

As mentioned above, the models were trained in two stages. In the initial partial fine-tuning stage, only the new final layers were trained using hyperparameters selected from these ranges: a learning rate from 1e-3 to 1e-1 (on a logarithmic scale), weight decay from 1e-6 to 1e-3 (on a logarithmic scale), dropout rate from 0.2 to 0.5, and a number of epochs from 10 to 50. In the subsequent full fine-tuning stage --- where all layers were updated --- the hyperparameter search was conducted over the following ranges: a learning rate from 1e-6 to 1e-4 (on a logarithmic scale), weight decay from 1e-6 to 1e-3 (on a logarithmic scale), and 10 to 50 epochs.

Each sampled hyperparameter combination defined an epoch cap. In inner CV, we trained each fold for up to the epoch cap and retained the checkpoint with the lowest validation loss. After selecting the best combination (lowest mean inner-CV loss), we set the outer-loop epoch cap to the maximum of the five best-checkpoint epochs plus 5, clipped to the selected configuration's cap. Outer models trained to this cap and predictions used the best-validation checkpoint within those epochs. We evaluated 30 hyperparameter combinations per stage.
\subsection{Estimators and Metrics}
\label{sec:org137732b}
\label{sec:sup-metrics}

With the optimal hyperparameters determined, in each outer repetition and fold iteration we trained a new model instance on the four training folds (group-A means) and evaluated it on the corresponding held-out test fold (group-B means). Only predictions from held-out test folds were used. We stored both raw and clipped predictions, but all reported metrics and analyses use the clipped (0–-100) predictions. Primary performance metric was mean absolute error (MAE), whereas root mean squared error (RMSE) and variance explained (\(R^2\)) are reported as secondary metrics. We complemented point estimates with 95\% confidence intervals (CIs) computed via the hierarchical bootstrap described below.
\subsubsection{Single-Model Estimator}
\label{sec:orgc4b9257}
For each model configuration and each outer repetition, we concatenated the test-set predictions from its five folds and computed MAE, RMSE, and \(R^2\) against the held-out targets (group-B image means) on this concatenated test set. We then averaged these repetition-level metrics across the 10 outer repetitions to obtain the final single-model estimates reported in Table \ref{tab:predictions} (columns MAE, RMSE, \(R^2\)). These estimates constitute our primary performance results, and they reflect the expected performance achieved by a single trained model instance.
\subsubsection{Cross-Split Ensembling}
\label{sec:orgbdd5de3}
To reduce variance via model averaging and potentially improve accuracy \autocite{breimanBaggingPredictors1996,hansenNeuralNetworkEnsembles1990}, we report ensembling as a secondary estimate. In this paper, \emph{ensembling} denotes cross-split averaging across outer repetitions: for each image, we averaged its 10 out-of-fold test predictions (one from each outer repetition) to obtain a per-image ensemble prediction, yielding strictly out-of-fold predictions (no in-sample predictions). We computed MAE\textsubscript{ens}, RMSE\textsubscript{ens}, and \(R_{ens}^2\) from these ensemble predictions against Group-B means. Because outer repetitions use different train/test partitions, ensemble members are trained on partially different training sets; across members, the union of training images is larger than for any single instance. This increased training coverage can make our cross-split ensembling more optimistic than standard same-split ensembling, which isolates model averaging without changing the train set. We therefore keep single-model results as primary and treat ensemble results as secondary.
\subsubsection{Uncertainty Estimation via Hierarchical Bootstrap}
\label{uncertainty-estimation}
We obtained 95\% confidence intervals via a hierarchical bootstrap applied to the aligned held-out image set (images present in all outer repetitions). For single-model estimates, each draw resampled outer repetitions with replacement and, within every resampled repetition, resampled images with replacement from that repetition's held-out set. MAE, RMSE, and \(R^2\) were recomputed per repetition on the resampled images and then averaged within the draw. For the ensemble mode, each draw first resampled outer repetitions with replacement and, for each image, formed an ensemble prediction by averaging the sampled repetitions' out-of-fold test predictions; it then resampled images with replacement and recomputed MAE, RMSE, and \(R^2\) once from these per-image ensemble predictions. The key distinction is that single-model results reflect averaging of repetition-level metrics, whereas ensemble results reflect metrics computed from per-image averaged predictions.

For both estimates, each draw uses a bootstrap sample of size 10 over outer repetitions (duplicates allowed), and a bootstrap sample of size 313 over images. We generated 5,000 draws per configuration with a fixed random seed, which for ensembling corresponds to 5,000 bootstrap ensembles of size 10. All metrics use clipped predictions (0--100) and Group-B image means on held-out folds.
\subsubsection{Upper Bound on Explainable Variance}
\label{sec:orga28aadb}

To estimate the upper bound of the models' predictive performance and to assess whether it is constrained by the number of ratings per image, we quantified inter-rater agreement and its stability as rater count increases. Specifically, we summarize the ceiling on explainable variance with \(\text{ICC(2,k)}\)  \autocite{revellePsychProceduresPsychological2024}, interpreted as the proportion of between-image variance reliably captured by the group mean. A stabilization of ICC across rater counts indicates a sufficient number of raters for the bound to be reliable and that group-level results are not limited by the number of ratings per image. This would also support the use of a single fixed participant split by suggesting limited sensitivity to split variance at the group level.

We computed \(\text{ICC(2,k)}\) with R's psych package \autocite{revellePsychProceduresPsychological2024}, suitable for two-way random effects and average measures. To examine stability across rater counts, we performed a bootstrap over raters: for each subsample size (9, 18, 27, 37, 46, 55, 64, and 74), we sampled participants without replacement from the full dataset and recomputed \(\text{ICC(2,k)}\), repeating this 100 times. Subsampling can induce missingness at the image level when, in a given draw, no sampled raters remain for a particular image; the R implementation accommodates this draw-induced missingness. 
\subsection{Learning Curve Analysis Details}
\label{sec:orgcd0d0ee}
\label{sec:sup-lc}

To quantify how performance improvements scale with dataset size, we fitted parametric learning curves to the averaged MAE, RMSE, and \(R^2\) scores across the seven dataset sizes. Besides the focus on partial fine-tuning, single-model estimates, and the number of outer repetitions (as described in Methods), all other settings matched the primary analysis. The training procedure for the complete dataset was already part of our primary analysis and was therefore not repeated. The results of these 15 repetitions were then averaged to obtain the final MAE, RMSE and \(R^2\) values for each dataset size.

For MAE and RMSE, we used exponential decay to a limit: \(y(n) = a \cdot exp(-b \cdot n) + c\). For \(R^2\), we used an exponential rise to a limit: \(y(n) = a \cdot (1 - exp(- b \cdot n)) + c\). Parameters (a, b, c) were estimated via weighted nonlinear least squares in Python (\texttt{scipy.optimize.curve\_fit}; \cite{virtanenSciPy10Fundamental2020}), using per-size standard errors (SEs) as weights (inverse-variance weighting, treating SEs as known; in SciPy: \texttt{sigma=SE, absolute\_sigma=True}). This accounts for heteroskedasticity across sizes (higher subsampling noise at smaller \(N\)).

The rate \(b\) captures how quickly the curve moves toward that plateau. We quantified convergence using the fitted rate \(b\) and the asymptotic plateau \(y_{\infty}\) (the limit of the fitted curve as dataset size \(n \to \infty\)). For decreasing metrics (MAE, RMSE), \(y_{\infty}=c\); for increasing metrics (\(R^2\)), \(y_{\infty}=c+a\). We summarize early gain and near-convergence using \(x_{50}\) and \(x_{95}\), defined as the dataset sizes required to reach 50\% and 95\% of the total improvement from the initial value to \(y_{\infty}\) (for the increasing form \(y(0)=c\); for the decreasing form \(y(0)=c+a\)). For the increasing form \(y(n)=c+a\,\big(1-\mathrm{e}^{-bn}\big)\), the fraction of improvement achieved after \(n\) images is \(1-\mathrm{e}^{-bn}\). Solving \(1-\mathrm{e}^{-bn}=p\) for \(n\) (with \(p\in[0,1]\)) gives

\begin{equation}\label{eq:lc}
x_p=\ln\!\left(\frac{1}{1-p}\right)/b,
\end{equation}

so \(x_{0.5}=\ln(2)/b\) and \(x_{0.95}=\ln(20)/b\) (reported as \(x_{50}\) and \(x_{95}\) in figures). For decreasing metrics (MAE, RMSE), “improvement” refers to reduction toward the asymptotic plateau, and the same \(x_p\) applies; the decreasing form is \(y(n)=c+a\,\mathrm{e}^{-bn}\). We report \(y_{\infty}\), \(x_{50}\), and \(x_{95}\).
\subsection{Attributions Details}
\label{sec:orgdfe9c92}
\label{sec:attributions-details}

As described in the main text (see \hyperref[attributions]{Attributions}), we used Grad-CAM to test whether model attributions concentrated on spider regions when predicting fear. Here we provide implementation details and the exact quantification procedure.

We computed Grad-CAM heatmaps using the pytorch-grad-cam library \autocite{jacobgilpytorchcam} for each of the four task-adapted base models (ResNet, ConvNeXtV2, DeiT, DINOv2). For CNNs, the target layer was the last feature stage that still preserved a spatial grid (the final residual block in ResNet and the last stage in ConvNeXtV2). For transformers, we used the output of the last transformer block, reshaped token activations to the 2D patch grid, dropped prefix tokens (e.g., class and register tokens), and then upsampled the resulting maps to the original image resolution with bilinear interpolation.

For each base model, we selected the configuration with the lowest single-model MAE in the primary analysis and reused the corresponding nested cross-validation splits and checkpoints (see \hyperref[model-evaluation]{Model Evaluation}). For every outer repetition and fold (10 repetitions \texttimes{} 5 folds), Grad-CAM heatmaps were generated only for the held-out test images in that fold. Because each image appears exactly once per outer repetition, this produced one heatmap per image per repetition. We then averaged these per-image heatmaps across the 10 outer repetitions to obtain a single, repetition-averaged heatmap for each image and model configuration.

Semantic segmentation masks for spiders were obtained using Label Studio (\href{http://labelstud.io}{\uline{labelstud.io}}). A trained researcher manually annotated spider pixels in all images that contained spiders, resulting in 281 annotated images (out of 313). Images without spiders were excluded from the attribution analysis.

Quantification was based on mean Grad-CAM activation levels inside and outside the spider masks. For image \(i\), we defined \(\mu_{in,i}\) as the mean Grad-CAM value over all pixels inside the mask and \(\mu_{out,i}\) as the mean over all pixels outside the mask. We used raw Grad-CAM scores without per-image rescaling and treated each image as one observation. The primary summary statistic was the paired difference

\begin{equation}\label{eq:delta-mu}
\Delta\mu_i = \mu_{in,i} - \mu_{out,i},
\end{equation}

with \(\Delta\mu_i > 0\) indicating higher activation inside the spider mask. We conducted a one-sided paired \(t\text{-test}\) on the \(\Delta\mu_i\) values across the 281 images to test whether mean activation inside the mask exceeded mean activation outside the mask. Effect sizes were quantified using Cohen's \(d\) for paired samples. Mask sizes varied substantially across images (approximately 5–80\% of pixels). We weighted each image equally, rather than by mask size, so that the analysis captured the average image-level tendency to focus on spiders instead of being driven disproportionately by images with very large or very small spider regions. Full per-model summaries and distributions of \(\Delta\mu_i\) are shown in Supplementary Results (Appendix \ref{sec:gc}).
\subsection{Feature Visualization Details}
\label{sec:orge22699d}
\label{sec:sup-fv}
We generated feature visualizations (FVs) with torch-lucent \autocite{kiatTorchlucent2020} to synthesize inputs that maximize the final scalar output (fear) of the task-adapted ResNet model trained on the full dataset (fully fine-tuned; lowest mean absolute error). We used Fourier parameterization (\texttt{fft=True}) with color decorrelation (\texttt{decorrelate=True}), \texttt{sd=0.001}, and standard augmentations (pad, jitter, random scale 0.8--1.2, small rotations). Adam (\texttt{lr=0.05}; \texttt{weight\_decay=1e-4}) ran for 512 steps. Identical settings applied to ConvNeXtV2 yielded high-frequency, iridescent patterns with low human interpretability; therefore, we report ResNet FVs and include ConvNeXtV2 attempts as negative results. Output clipping was disabled during FV. We generated 500 synthetic images per base model and show 9 representatives (selected by highest final activation) in Figure \ref{fig:xai}b. The full set of FV outputs is available at the OSF repository.
\subsection{Error Analysis Details}
\label{sec:org98d17f4}
\label{sec:error-details}

This section expands on the main text (see \hyperref[error-analysis]{Error Analysis}) by providing additional details, such as variable definitions, predictor specifications, distance computations, transformations, the model equation, and reporting conventions. As mentioned in the main text, the dependent variable for the mixed-effects analysis was the log-transformed absolute prediction error. For image \(i\) in outer repetition \(r\), with prediction \(\hat{y}_{r,i}\) and evaluation target \(y_{i}\) (Group-B mean), we defined the absolute error as \(AE_{r,i}=\lvert \hat{y}_{r,i}-y_i \rvert\) and the log-transformed error as \(L_{r,i}=\log(AE_{r,i}+10^{-3})\) to handle zeros. Image-level variability in \(L_{r,i}\) was captured by the random intercepts.

Predictors comprised four components. First, image categories coded following \textcite{karnerVisualAttributesSpiders2024} included spider type (no spider, real, artificial) and general attributes (environment: nature, civilization, human contact; color: color, black/white; cobweb presence: present, absent). Spider-only attributes were gated by multiplying with a binary spider-present indicator and included subjective distance (close, distant), subjective size (small, middle, large), texture (smooth, hairy), eyes (non-visible, visible), eating prey (not eating, eating), number of spiders (one, two or more), perspective (side/front, top/bottom), and prominent legs (no, yes). Categorical predictors were effect-coded (sum-to-zero) \autocite{singmannIntroductionMixedModels2019}, so coefficients represent deviations from the grand mean across all levels. Second, fear ratings were included as the image mean \(z_{\mathrm{fear}}\) and its quadratic term \(z_{\mathrm{fear}}^2\) to model the possibility that errors are higher at the extremes. Third, representational distance \(z_{\mathrm{dist}}\) was computed per base model from L2-normalized penultimate-layer embeddings (also called \emph{pre-logits}; CNNs: global-average-pooled features, transformers: class [CLS] token at the last block). For each held-out image, we averaged its cosine distances to all training images within the same outer fold, using only that fold's training split to avoid leakage. We included base-model fixed effects in the regression to absorb remaining level differences across representation spaces. Fourth, target shift \(z_{\mathrm{shift}}\) was \(\log(1{+}|\mathbf{B}{-}\mathbf{A}|)\), where \(|\mathbf{B}{-}\mathbf{A}|\) is the absolute difference between Group-B (held-out) and Group-A (training) image means. We used the absolute difference to capture disagreement magnitude irrespective of direction and applied \(\log(1{+}\cdot)\) because raw \(|\mathbf{B}{-}\mathbf{A}|\) is heavily right-skewed with many near-zero cases. The log transformation compresses the tail, permits zeros, and reflects diminishing returns whereby small shifts near zero have larger incremental impact than equally sized increases in already large shifts. All continuous predictors (\(z_{\mathrm{fear}}\), \(z_{\mathrm{fear}}^2\), \(z_{\mathrm{shift}}\), and \(z_{\mathrm{dist}}\)) were z-standardized across the pooled analysis dataset spanning all base models and repetitions.

We fit a single full specification that includes all predictor blocks:

\begin{equation}\label{eq:error-model}
\log|AE| \sim z_{\mathrm{dist}} + z_{\mathrm{fear}} + z_{\mathrm{fear}}^2 + \text{categories} + z_{\mathrm{shift}} + \text{base-model fixed effects}.
\end{equation}

Coefficients are reported on the log scale; for interpretation, a coefficient \(\beta\) corresponds to an approximate percent change in absolute error of \((\exp(\beta)-1)\times 100\). To account for multiple testing across predictors, we applied Benjamini–Hochberg false discovery rate (FDR) correction \autocite{benjaminiControllingFalseDiscovery1995} to the two-sided \(p\text{-values}\) of all fixed effects. As descriptive diagnostics, we summarize signed errors (\(\hat{y}-y\)) and rater-group differences with means, standard deviations, and 95\% CIs and show the corresponding histograms in Supplementary Results in Appendix \ref{sec:err} (Figures \ref{fig:signed_error_hist_per_model} and \ref{fig:diff_ab_hist}).
\section{Supplementary Results}
\label{sec:orgcd91d50}
\label{sec:sup-results}
\subsection{Data Descriptives}
\label{sec:org4cd0ae3}
\label{sec:data-distributions}

\begin{figure}[htbp]
  \centering \includegraphics[width=0.8\linewidth]{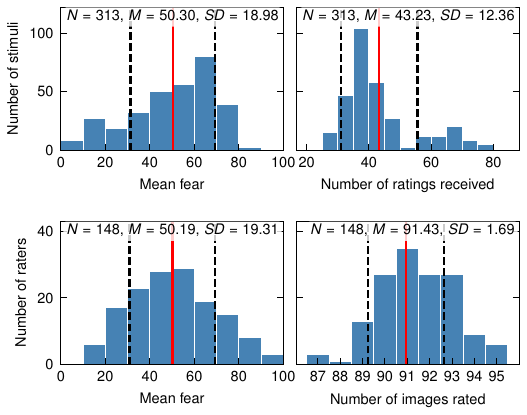} \caption{\label{fig:data-distributions-all}Distributions of Images and Participants Based on Mean Fear Ratings and Number of Ratings Provided} \par\footnotesize\textit{Note}. The top row shows the distribution of images by their mean fear ratings (left) and the number of ratings they received (right). The bottom row shows the distribution of participants by their mean fear ratings (left) and the number of ratings they provided (right).
\end{figure}

\begin{figure}[htbp]
  \centering \includegraphics[width=0.8\linewidth]{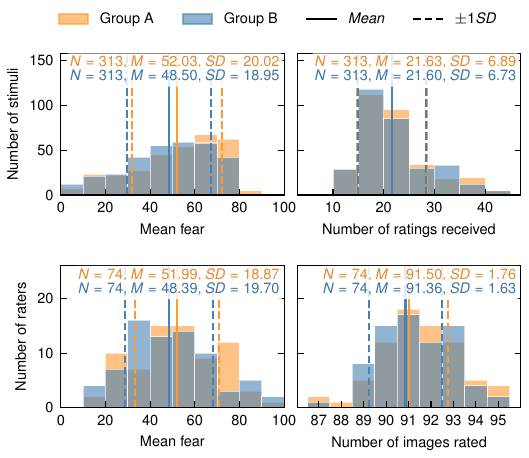}
  \caption{\label{fig:data-distributions-ab}Overlay of Distributions for Groups A and B Based on Mean Fear Ratings and Number of Ratings}
  \par\footnotesize\textit{Note}. Bars are overlaid: Group A (orange) and Group B (blue). Solid vertical lines indicate group means; dashed lines indicate $\pm$1 SD. Top row: images by mean fear (left) and number of ratings received (right). Bottom row: participants by mean fear (left) and number of images rated (right).
\end{figure}

\FloatBarrier
\clearpage
\subsection{Predictive Performance}
\label{sec:org6a9c275}
\label{sec:predictive-performance}

\begin{figure*}[htb]
  \centering \includegraphics[width=\textwidth]{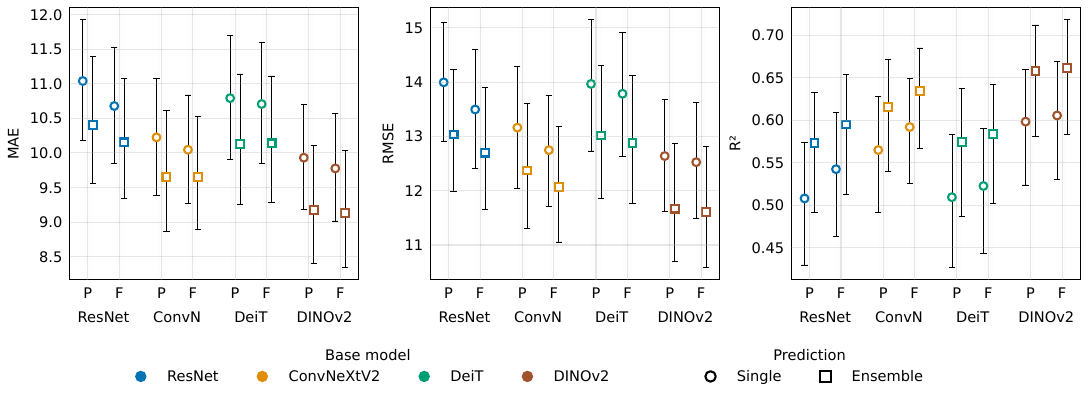}
  \caption{\label{fig:performance-bars}Predictive Performance (MAE, RMSE, $R^2$) by Base Model, Fine-Tuning, and Prediction Mode}
  \par\footnotesize\textit{Note}. Points show point estimates; vertical error bars are 95\% hierarchical bootstrap confidence intervals. Groups are by base model; P (Partial) and F (Full) denote the fine-tuning strategies. Marker shape indicates prediction mode (circle = single, square = ensemble); color encodes base model. Estimates are computed on held-out test images; see Methods for details. Full numeric details appear in Table~\ref{tab:predictions}.
\end{figure*}

\begin{table*}[t]
\centering
\caption{Predictive performance of ResNet-, ConvNeXtV2-, DeiT-, and DINOv2-based models}
\label{tab:predictions}
\begin{threeparttable}
\fontsize{8}{10}\selectfont
\begin{tabularx}{\textwidth}{llllS[table-format=2.2]rS[table-format=2.2]rS[table-format=1.2]r}
\toprule
\addlinespace[0.3em]
 &  &  & Metric & \multicolumn{2}{c}{${MAE}$} & \multicolumn{2}{c}{${RMSE}$} & \multicolumn{2}{c}{${R^{2}}$} \\
\cmidrule(lr){5-6}
\cmidrule(lr){7-8}
\cmidrule(lr){9-10}
\cmidrule(lr){5-10}
 &  &  &  & \multicolumn{1}{c}{Est} & 95\% CI & \multicolumn{1}{c}{Est} & 95\% CI & \multicolumn{1}{c}{Est} & 95\% CI \\
Base model & Fine-tuning & Train time & Prediction &  &  &  &  &  &  \\
\midrule
ResNet & Partial & 285 & Single & 11.04 & [10.18, 11.92] & 14.00 & [12.91, 15.11] & 0.51 & [0.43, 0.57] \\
 &  &  & Ensemble & 10.40 & [9.56, 11.39] & 13.04 & [11.99, 14.24] & 0.57 & [0.49, 0.63] \\
 & Full & 881 & Single & 10.68 & [9.85, 11.53] & 13.50 & [12.41, 14.61] & 0.54 & [0.46, 0.61] \\
 &  &  & Ensemble & 10.15 & [9.34, 11.08] & 12.70 & [11.65, 13.90] & 0.60 & [0.51, 0.65] \\
ConvNeXtV2 & Partial & 347 & Single & 10.23 & [9.39, 11.08] & 13.16 & [12.04, 14.29] & 0.57 & [0.49, 0.63] \\
 &  &  & Ensemble & 9.65 & [8.86, 10.62] & 12.37 & [11.30, 13.61] & 0.62 & [0.54, 0.67] \\
 & Full & 1267 & Single & 10.05 & [9.26, 10.84] & 12.75 & [11.71, 13.76] & 0.59 & [0.53, 0.65] \\
 &  &  & Ensemble & 9.65 & [8.89, 10.53] & 12.07 & [11.06, 13.18] & 0.63 & [0.57, 0.69] \\
DeiT & Partial & 470 & Single & 10.79 & [9.91, 11.70] & 13.97 & [12.73, 15.15] & 0.51 & [0.43, 0.58] \\
 &  &  & Ensemble & 10.13 & [9.25, 11.14] & 13.02 & [11.85, 14.31] & 0.57 & [0.49, 0.64] \\
 & Full & 1276 & Single & 10.71 & [9.85, 11.60] & 13.79 & [12.64, 14.91] & 0.52 & [0.44, 0.59] \\
 &  &  & Ensemble & 10.14 & [9.29, 11.11] & 12.88 & [11.76, 14.13] & 0.58 & [0.50, 0.64] \\
DINOv2 & Partial & 356 & Single & 9.93 & [9.18, 10.70] & 12.64 & [11.62, 13.69] & 0.60 & [0.52, 0.66] \\
 &  &  & Ensemble & 9.17 & [8.41, 10.11] & 11.67 & [10.69, 12.86] & 0.66 & [0.58, 0.71] \\
 & Full & 1023 & Single & 9.78 & [9.02, 10.57] & 12.53 & [11.49, 13.63] & 0.61 & [0.53, 0.67] \\
 &  &  & Ensemble & 9.13 & [8.34, 10.03] & 11.61 & [10.59, 12.82] & 0.66 & [0.58, 0.72] \\
\bottomrule
\end{tabularx}
\begin{tablenotes}
\item \textit{Note.} MAE = mean absolute error; RMSE = root mean squared error; $R^{2}$ = coefficient of determination; estimates are on held-out test images; CIs are 95\% hierarchical bootstrap (single: resample one outer repetition and images; ensemble: resample repetitions and images); Train time = total training time (min) per outer repetition.
\end{tablenotes}
\end{threeparttable}
\end{table*}

\begin{figure*}[htb]
  \centering \includegraphics[width=\textwidth]{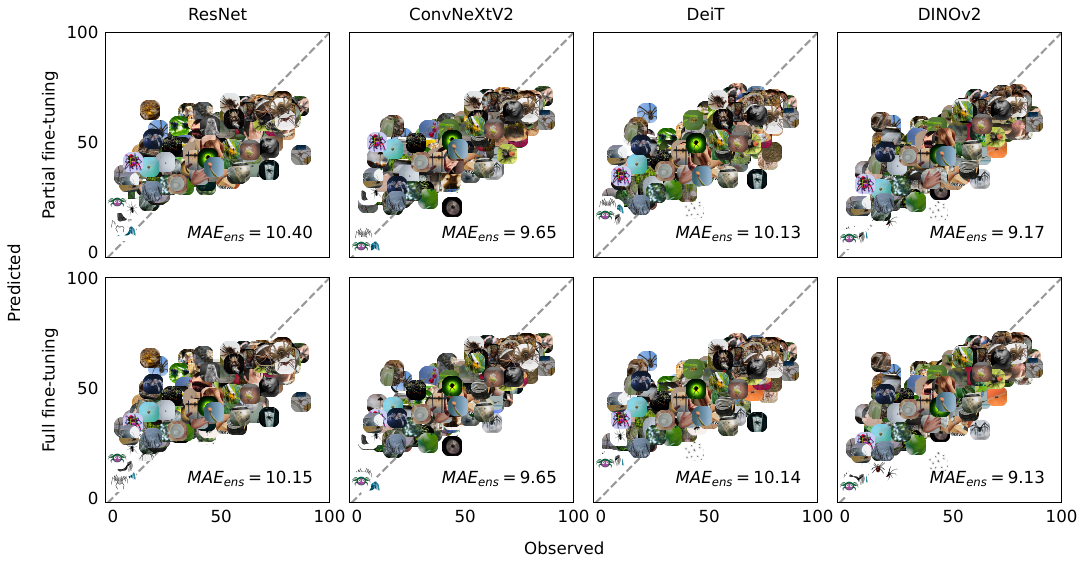}
  \caption{\label{fig:predictions} Observed Fear vs. Ensemble Predictions}
\par\footnotesize\textit{Note}. The predicted positions of each image are the ensemble predictions (averaged across outer repetitions), the mean absolute error ($MAE_{ens}$) is based on these predictions; the dashed identity line shows the ideal prediction.
\end{figure*}

\FloatBarrier
\clearpage
\subsection{Individual- vs. Group-Level Errors}
\label{sec:org171747f}
\label{sec:individual-errors}

For completeness, we compared errors when predicting individual fear ratings versus image-level group means. In our data, the MAE for individual ratings was 25.76 for the best single model and 25.56 for the ensemble, compared with 9.78 and 9.13 for group-level means. As expected, predicting single-trial responses is harder than predicting image-level averages, reflecting additional variability across individuals and occasions.

\FloatBarrier
\clearpage
\subsection{Upper Bound Estimation}
\label{sec:org551564c}
\label{sec:icc}

\begin{figure}[htpb]
\centering
\includegraphics[width=.8\linewidth]{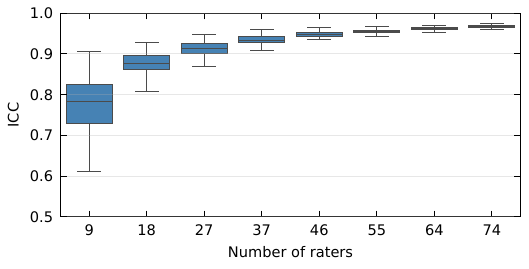}
\caption{\label{fig:icc}Intraclass Correlation Coefficient by Number of Raters}
\par\footnotesize\textit{Note}. Distribution of $\text{ICC(2,k)}$ over 100 subsampling repetitions for each subsample size (number of raters). Boxes show the median (center line) and interquartile range (IQR); whiskers extend to $1.5 \times \text{IQR}$.
\end{figure}

\FloatBarrier
\clearpage
\subsection{Learning Curve Analysis}
\label{sec:orgc4beee3}
\label{sec:learning-curve-analysis}

\begin{figure*}[htb]
  \centering \includegraphics[width=\textwidth]{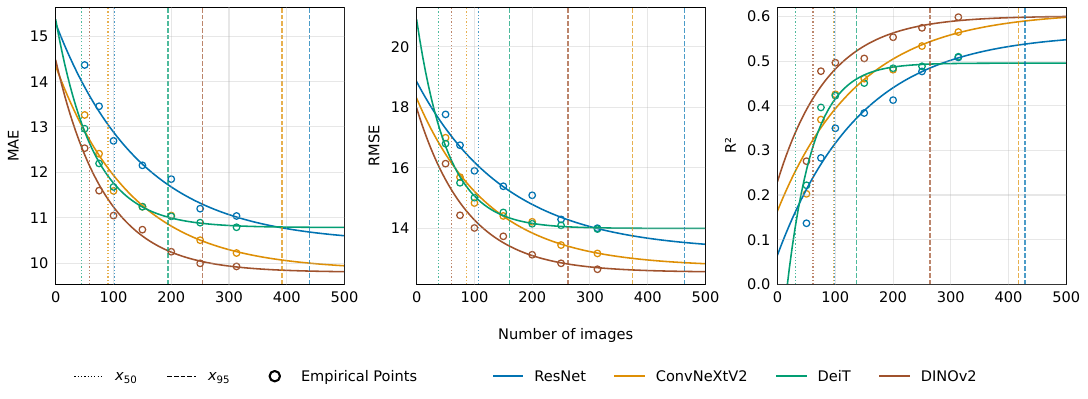}
  \caption{\label{fig:learning-curves} Learning Curves for Mean Absolute Error (MAE), Root Mean Squared Error (RMSE), and Variance Explained ($R^2$)}
\par\footnotesize\textit{Note}. Solid lines show fitted exponential learning curves; circles denote empirical scores at each data-set size. Vertical lines mark $x_{50}$ (dotted) and $x_{95}$ (dashed), color-matched to models. See Table~\ref{tab:learning-curve-convergence} for per-model $y_{\infty}$, $x_{50}$, and $x_{95}$ values.
\end{figure*}

\begin{table*}[h!]
\centering
\caption{Convergence Metrics From Fitted Learning Curves}
\label{tab:learning-curve-convergence}
\begin{threeparttable}
\fontsize{10}{12}\selectfont
\begin{tabular}{lrrrrrrrrr}
\toprule
\addlinespace[0.3em]
 & \multicolumn{3}{c}{MAE} & \multicolumn{3}{c}{RMSE} & \multicolumn{3}{c}{$R^2$} \\
\cmidrule(lr){2-4}
\cmidrule(lr){5-7}
\cmidrule(lr){8-10}
Model & $y_{\infty}$ & $x_{50}$ & $x_{95}$ & $y_{\infty}$ & $x_{50}$ & $x_{95}$ & $y_{\infty}$ & $x_{50}$ & $x_{95}$ \\
\midrule
ResNet & 10.45 & 101.60 & 439.12 & 13.24 & 107.43 & 464.29 & 0.56 & 99.12 & 428.39 \\
ConvNeXtV2 & 9.85 & 90.63 & 391.68 & 12.72 & 86.32 & 373.07 & 0.61 & 96.37 & 416.49 \\
DeiT & 10.79 & 44.98 & 194.40 & 13.99 & 37.25 & 161.00 & 0.50 & 31.53 & 136.25 \\
DINOv2 & 9.80 & 58.93 & 254.71 & 12.54 & 60.63 & 262.02 & 0.60 & 61.06 & 263.88 \\
\bottomrule
\end{tabular}
\begin{tablenotes}
\item \textit{Note.} Plateau $y_{\infty}$ is defined as $c$ for decreasing metrics (MAE, RMSE) and as $c+a$ for increasing metrics ($R^2$). Convergence points are $x_{50}=\ln(2)/b$ and $x_{95}=\ln(20)/b$.
\end{tablenotes}
\end{threeparttable}
\end{table*}

\begin{table*}[htbp]
\centering
\caption{Empirical Learning–Curve Values}
\label{tab:empirical-learning-curves}
\begin{threeparttable}
\fontsize{8}{10}\selectfont
\begin{tabular}{lrrrrrrrrrrrr}
\toprule
\addlinespace[0.3em]
Data Size & \multicolumn{4}{c}{MAE} & \multicolumn{4}{c}{RMSE} & \multicolumn{4}{c}{$R^2$} \\
\cmidrule(lr){1-1}
\cmidrule(lr){2-5}
\cmidrule(lr){6-9}
\cmidrule(lr){10-13}
 & ResNet & ConvN & DeiT & DINOv2 & ResNet & ConvN & DeiT & DINOv2 & ResNet & ConvN & DeiT & DINOv2 \\
\midrule
50 & 14.36 & 13.26 & 12.96 & 12.53 & 17.76 & 16.99 & 16.79 & 16.13 & 0.14 & 0.20 & 0.22 & 0.28 \\
75 & 13.45 & 12.41 & 12.20 & 11.60 & 16.74 & 15.68 & 15.50 & 14.42 & 0.28 & 0.37 & 0.40 & 0.48 \\
100 & 12.69 & 11.59 & 11.68 & 11.05 & 15.89 & 14.83 & 15.01 & 14.00 & 0.35 & 0.43 & 0.42 & 0.50 \\
150 & 12.15 & 11.24 & 11.24 & 10.74 & 15.38 & 14.40 & 14.52 & 13.73 & 0.38 & 0.46 & 0.45 & 0.51 \\
200 & 11.85 & 11.05 & 11.03 & 10.25 & 15.08 & 14.21 & 14.14 & 13.12 & 0.41 & 0.48 & 0.48 & 0.55 \\
250 & 11.20 & 10.51 & 10.89 & 10.00 & 14.29 & 13.44 & 14.09 & 12.84 & 0.48 & 0.53 & 0.49 & 0.57 \\
313 & 11.04 & 10.23 & 10.79 & 9.93 & 14.00 & 13.16 & 13.97 & 12.64 & 0.51 & 0.57 & 0.51 & 0.60 \\
\bottomrule
\end{tabular}
\begin{tablenotes}
\item \textit{Note.} MAE = Mean Absolute Error; RMSE = Root Mean Squared Error; $R^2$ = Coefficient of Determination (Variance Explained).The numbers are empirical validation scores obtained at each data set size and are the data points used to fit the learning curves in Figure~\ref{fig:learning-curves}. ConvNeXtV2 is abbreviated as 'ConvN'.
\end{tablenotes}
\end{threeparttable}
\end{table*}

\begin{table*}[htbp]
\centering
\caption{Parameters (a, b, c) of the Fitted Learning-Curve Functions}
\label{tab:learning-curve-params}
\begin{threeparttable}
\fontsize{10}{12}\selectfont
\begin{tabular}{lrrrrrrrrrrrr}
\toprule
\addlinespace[0.3em]
Metric & \multicolumn{3}{c}{ResNet} & \multicolumn{3}{c}{ConvNeXtV2} & \multicolumn{3}{c}{DeiT} & \multicolumn{3}{c}{DINOv2} \\
\cmidrule(lr){1-1}
\cmidrule(lr){2-4}
\cmidrule(lr){5-7}
\cmidrule(lr){8-10}
\cmidrule(lr){11-13}
 & a & b & c & a & b & c & a & b & c & a & b & c \\
\midrule
MAE & 4.827 & 0.007 & 10.446 & 4.487 & 0.008 & 9.847 & 4.565 & 0.015 & 10.787 & 4.659 & 0.012 & 9.802 \\
RMSE & 5.595 & 0.006 & 13.243 & 5.564 & 0.008 & 12.722 & 6.889 & 0.019 & 13.989 & 5.422 & 0.011 & 12.541 \\
R² & 0.496 & 0.007 & 0.066 & 0.445 & 0.007 & 0.165 & 0.718 & 0.022 & -0.223 & 0.370 & 0.011 & 0.231 \\
\bottomrule
\end{tabular}
\begin{tablenotes}
\item \textit{Note.} Parameters a, b, c are those of the functions that generated the learning-curve lines in Figure~\ref{fig:learning-curves}. For coefficient of determination ($R^2$) the fitted form is an inverse exponential $y(n) = a \cdot (1 - exp(-b n)) + c$. For error metrics (MAE, RMSE) the fitted form is an exponential decay $y(n) = a \cdot exp(-b n) + c$. Parameter estimates are rounded to 3 decimal places. All subsequent calculations were based on the full-precision (i.e., unrounded) parameters.
\end{tablenotes}
\end{threeparttable}
\end{table*}

\FloatBarrier
\clearpage
\subsection{Grad-CAM Attributions}
\label{sec:org1b54140}
\label{sec:gc}

\begin{figure*}[htb]
  \centering \includegraphics[width=.8\textwidth]{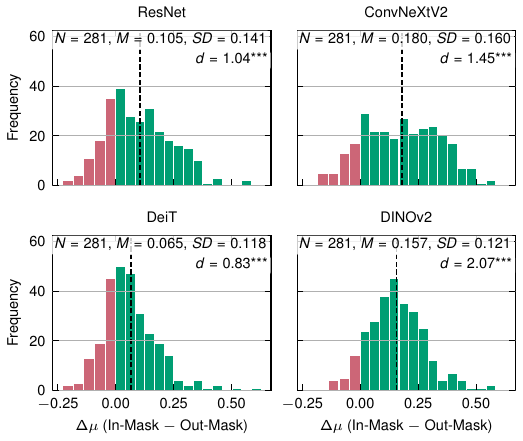}
  \caption{\label{fig:gradcam-difference-distribution} Differences in Mean Grad-CAM Attributions Between In-Mask and Out-Mask}
\par\footnotesize\textit{Note}. For image $i$, $\mu_{in,i}$ and $\mu_{out,i}$ are mean Grad-CAM activations inside/outside the mask; $\Delta\mu_i = \mu_{in,i} - \mu_{out,i}$. Green areas mark the proportion of images with higher Grad-CAM activation inside spider masks than outside; red marks the opposite. Asterisks denote paired one-sided $t$-tests (inside $>$ outside): * $p<0.05$; ** $p<0.01$; *** $p<0.001$; $d$ is effect size quantified with Cohen's $d$.\end{figure*}

\begin{figure*}[htp]
  \centering \includegraphics[width=\textwidth]{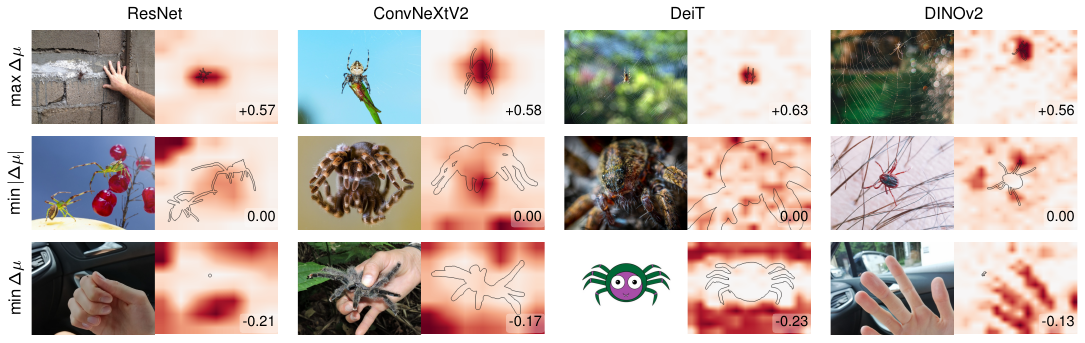}
  \caption{\label{fig:gradcam} Grad-CAM Examples per Base Model}
\par\footnotesize\textit{Note}. Figure shows Grad-CAM heatmaps for held-out spider images; $\Delta\mu = \mu_{in} - \mu_{out}$ (mean activation inside minus outside the annotated spider mask). Positive $\mu$ indicates greater focus on spider regions; negative indicates focus elsewhere. For each base model, the images with the largest positive activation difference (first row), the smallest absolute activation difference (middle row), and the largest negative activation difference (last row) are shown. The first row has more focus on the spider, the middle row has approximately equal focus on the spider and background, and the last row has more focus on the background.\end{figure*}

\FloatBarrier
\clearpage
\subsection{Feature Visualization}
\label{sec:orga7b1280}
\label{sec:fv}
\subsubsection{ResNet}
\label{sec:org3532255}

\begin{figure}[htbp]
\centering
\includegraphics[width=0.8\linewidth]{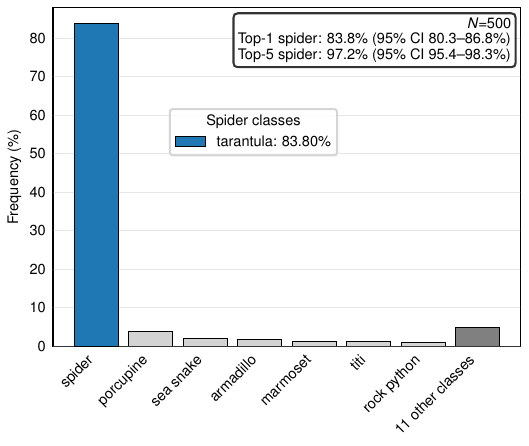}
\caption{\label{fig:fv-classes-resnet}ImageNet-1K Classes for Feature Visualizations Predicted by an Unadapted ResNet Base Model}
\end{figure}

\begin{figure}[htbp]
\centering
\includegraphics[width=0.8\linewidth]{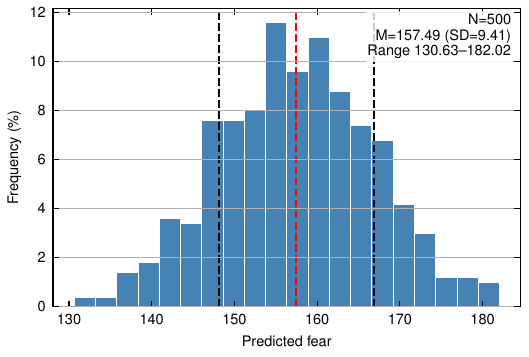}
\caption{\label{fig:fv-fear-resnet}Distribution of Predicted Fear for Feature Visualizations of the Task-Adapted ResNet Model}
\end{figure}

Figure \ref{fig:fv-fear-resnet} illustrates qualitative and quantitative FV outcomes (\(N=\) 500 FV images). An unadapted ResNet-ImageNet classifier assigned a spider synset (black and gold garden spider, barn spider, garden spider, black widow, tarantula, wolf spider, spider web) as top-1 for 83.8\% (95\% CI 80.3\%–86.8\%) (notably, all FV images classified as ``spider'' were assigned the specific class ``tarantula'') and as top-5 for 97.2\% (95\% CI 95.4\%–98.3\%). The mean spider probability mass was 0.572 (95\% CI 0.545–0.599). Predicted fear values from the task-adapted model (clipping disabled) had mean 157.49 (\(SD=\) 9.41), range 130.63–182.02.
See Figures \ref{fig:fv-classes-resnet} and Table \ref{tab:fv-summary}. Examples and the full set of FV outputs are available at the OSF repository.

\begin{table*}[p]
\centering
\caption{Summary of Feature Visualization Sanity Checks (N images per model)}
\label{tab:fv-summary}
\begin{threeparttable}
\fontsize{9}{11}\selectfont
\begin{tabular}{lrrrrr}
\toprule
\addlinespace[0.3em]
Model & N & \shortstack{Spider top-1\\(95\% CI)} & \shortstack{Spider top-5\\(95\% CI)} & \shortstack{Spider prob. mass\\(mean; 95\% CI)} & \shortstack{Fear\\(mean $\pm$ SD; min--max)} \\
\midrule
ResNet & 500 & 83.8\% (80.3–86.8\%) & 97.2\% (95.4–98.3\%) & 0.572 (0.545–0.599) & 157.49 $\pm$ 9.41 (130.63–182.02) \\
ConvNeXtV2 & 500 & 13.0\% (10.3–16.2\%) & 33.0\% (29.0–37.2\%) & 0.064 (0.054–0.074) & 101.12 $\pm$ 7.05 (90.85–119.36) \\
\bottomrule
\end{tabular}

\begin{tablenotes}
\item \textit{Note.} “Spider” synsets = {black and gold garden spider, barn spider, garden spider, black widow, tarantula, wolf spider, spider web}. Classifier = unadapted ImageNet-1K backbone with native preprocessing. Spider probability mass is the sum of softmax probabilities over the spider synsets. Task-adapted model clipping was disabled during FV.
\end{tablenotes}
\end{threeparttable}
\end{table*}
\subsubsection{ConvNeXtV2}
\label{sec:org24d2ba9}

\begin{figure}[htbp]
\centering
\includegraphics[width=0.8\linewidth]{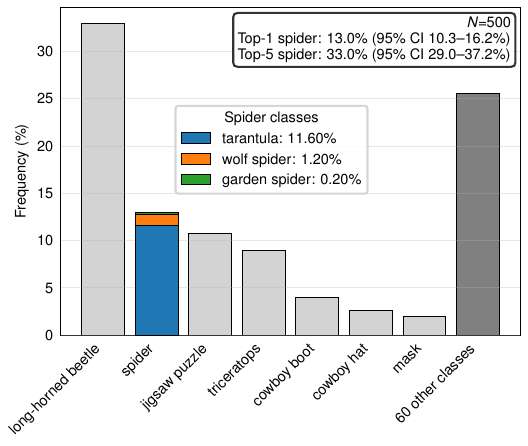}
\caption{\label{fig:fv-classes-convnextv2}ImageNet-1K Classes for Feature Visualizations Predicted by an Unadapted ConvNeXtV2 Base Model}
\end{figure}

\begin{figure}[htbp]
\centering
\includegraphics[width=0.8\linewidth]{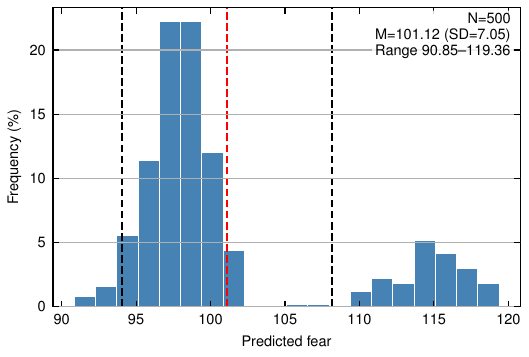}
\caption{\label{fig:fv-fear-convnextv2}Distribution of Predicted Fear for 500 Feature Visualizations of the Task-Adapted ConvNeXtV2 Model}
\end{figure}

Under identical FV settings, ConvNeXtV2 produced iridescent patterns lacking coherent object structure (\(N=\) 500 FV images); top-1 spider 13.0\% (95\% CI 10.3\%–16.2\%), top-5 spider 33.0\% (95\% CI 29.0\%–37.2\%); mean spider probability mass 0.064 (95\% CI 0.054–0.074). Predicted fear mean was 101.12 (\(SD=\) 7.05), range 90.85–119.36. See Figures \ref{fig:fv-classes-convnextv2}, \ref{fig:fv-fear-convnextv2} and Table \ref{tab:fv-summary}. Examples and the full set of FV outputs are available at the OSF repository.

\begin{figure}[tbp]
  \centering \includegraphics[width=0.8\linewidth]{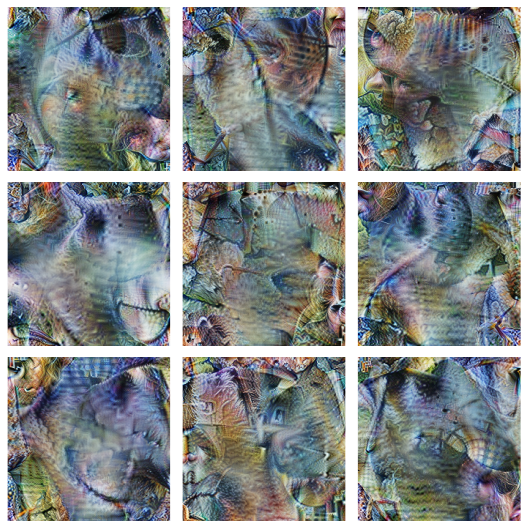}
  \caption{\label{fig:fv-grid-convnextv2} ConvNeXtV2 Feature Visualizations for the Final Output Node}
\par\footnotesize\textit{Note}.
The figure shows nine synthetic example images that maximally activate the task-adapted model's fear rating predictions, revealing the learned features that the model associates with high fear.\end{figure}

\FloatBarrier
\clearpage
\subsection{Error Analysis}
\label{sec:org0cfd7ca}
\label{sec:err}

This appendix complements the main text by providing the full coefficient table and diagnostic figures for the exploratory error analysis.

\begin{itemize}
\item Table \ref{tab:mixed-effects-full} lists all fixed effects for the full specification. Coefficients are on the log scale; for interpretation as percent change in absolute error use \((\exp(\beta) - 1) \times 100\). The table reports fixed-effect estimates from a linear mixed-effects model with random intercepts for images; standard errors are based on large-sample normal approximations and \(p\text{-values}\) are false discovery rate (FDR)-adjusted.
\end{itemize}

\begin{figure*}[t]
  \centering \includegraphics[width=\textwidth]{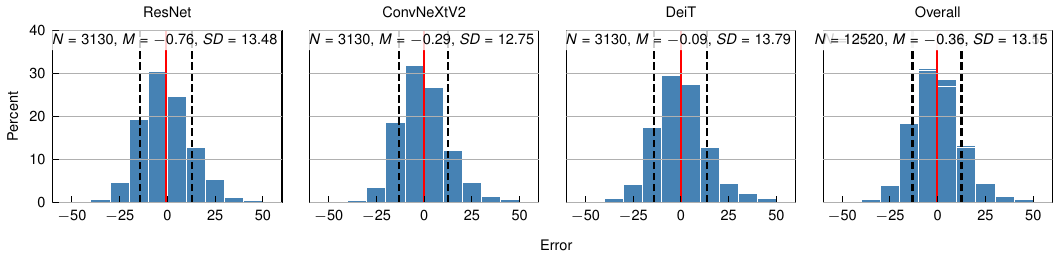}
  \caption{\label{fig:signed_error_hist_per_model} Signed Error Histograms by Base Model and Overall}
\par\footnotesize\textit{Note}. Percent histograms computed from single held-out predictions. For each base model, the fine-tuning strategy with the best predictive performance (from the primary analysis) was used. For all base models, this was the "full fine-tuning" strategy. Bins -60 to 60 in steps of 10; vertical grid aids comparison across panels.
\end{figure*}

\begin{figure}[htbp]
  \centering \includegraphics[width=.8\linewidth]{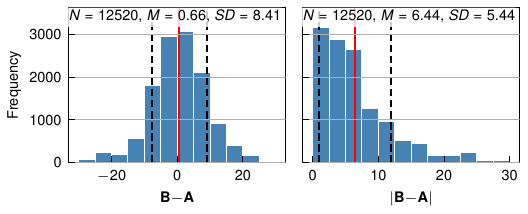}
  \caption{\label{fig:diff_ab_hist} Signed and Absolute Differences Between Group B and Group A Image Means}
  \par\footnotesize\textit{Note}.
Left: Signed difference ($\mathbf{B}{-}\mathbf{A}$). Right: The absolute difference ($\vert\mathbf{B}{-}\mathbf{A}\vert$) corresponds to the predictor $\mathbf{z_{\mathrm{shift}}}$ (log1p, z-standardized).

\end{figure}

\begin{itemize}
\item Figure \ref{fig:signed_error_hist_per_model} shows the distributions of signed errors \((\hat{y} - y)\) across single held-out predictions, with mean and SD.

\item Figure \ref{fig:diff_ab_hist} displays the distributions of \(\mathbf{B}-\mathbf{A}\) and \(|\mathbf{B}-\mathbf{A}|\) (signed and absolute differences between Group-B and Group-A image means).
\end{itemize}

\begin{longtable}{llS[table-format=+1.3]S[table-format=1.3]S[table-format=+2.3]ll}
\caption{All Fixed Effects On Log Absolute Error} \label{tab:mixed-effects-full} \\
\toprule
\addlinespace[0.3em]
Term & \multicolumn{1}{c}{Beta} & \multicolumn{1}{c}{SE} & \multicolumn{1}{c}{z} & $p_\mathrm{FDR}$ & \multicolumn{1}{c}{95\% CI} \\
\midrule
\endfirsthead
\caption[]{All Fixed Effects On Log Absolute Error} \\
\toprule
\addlinespace[0.3em]
Term & \multicolumn{1}{c}{Beta} & \multicolumn{1}{c}{SE} & \multicolumn{1}{c}{z} & $p_\mathrm{FDR}$ & \multicolumn{1}{c}{95\% CI} \\
\midrule
\endhead
\midrule
\multicolumn{6}{r}{Continued on next page} \\
\midrule
\endfoot
\bottomrule
\endlastfoot
Intercept & 1.761 & 0.140 & 12.615 & <.001 & [1.488, 2.035] \\
Held-out fear² (z) & 0.344 & 0.038 & 9.008 & <.001 & [0.269, 0.419] \\
Model: DINOv2 & -0.068 & 0.013 & -5.196 & <.001 & [-0.094, -0.042] \\
Model: ResNet & 0.052 & 0.014 & 3.715 & 0.002 & [0.025, 0.079] \\
Group mean shift |B-A| (log1p, z) & 0.114 & 0.033 & 3.391 & 0.006 & [0.048, 0.179] \\
Environment: civilization & 0.224 & 0.072 & 3.100 & 0.014 & [0.082, 0.365] \\
Model: DeiT & 0.033 & 0.013 & 2.571 & 0.059 & [0.008, 0.058] \\
Cobweb: present vs absent & -0.215 & 0.087 & -2.482 & 0.064 & [-0.386, -0.045] \\
Environment: not evaluable & -0.195 & 0.081 & -2.412 & 0.064 & [-0.353, -0.037] \\
Distance: distant (spider-present) & 0.118 & 0.050 & 2.362 & 0.064 & [0.020, 0.216] \\
Distance: close (spider-present) & -0.118 & 0.050 & -2.362 & 0.064 & [-0.216, -0.020] \\
Held-out fear (z) & 0.143 & 0.063 & 2.276 & 0.073 & [0.020, 0.266] \\
Texture: hairy (spider-present) & -0.087 & 0.047 & -1.831 & 0.181 & [-0.179, 0.006] \\
Texture: smooth (spider-present) & 0.087 & 0.047 & 1.831 & 0.181 & [-0.006, 0.179] \\
Prominent legs: no (spider-present) & 0.064 & 0.040 & 1.596 & 0.258 & [-0.015, 0.142] \\
Prominent legs: yes (spider-present) & -0.064 & 0.040 & -1.596 & 0.258 & [-0.142, 0.015] \\
Model: ConvNeXtV2 & -0.017 & 0.013 & -1.277 & 0.441 & [-0.042, 0.009] \\
Embedding distance (z) & -0.012 & 0.011 & -1.119 & 0.525 & [-0.034, 0.009] \\
Spider type: artificial & 0.120 & 0.109 & 1.103 & 0.525 & [-0.093, 0.333] \\
Perspective: side/front (spider-present) & -0.033 & 0.041 & -0.790 & 0.752 & [-0.113, 0.048] \\
Perspective: top/bottom (spider-present) & 0.033 & 0.041 & 0.790 & 0.752 & [-0.048, 0.113] \\
Eating prey: eating (spider-present) & 0.038 & 0.065 & 0.589 & 0.797 & [-0.089, 0.165] \\
Eating prey: not eating (spider-present) & -0.038 & 0.065 & -0.589 & 0.797 & [-0.165, 0.089] \\
Spider type: none & -0.066 & 0.117 & -0.560 & 0.797 & [-0.296, 0.164] \\
Spider type: real & -0.054 & 0.103 & -0.524 & 0.797 & [-0.256, 0.148] \\
Size: large (spider-present) & -0.037 & 0.071 & -0.521 & 0.797 & [-0.175, 0.102] \\
Number of spiders: one (spider-present) & -0.033 & 0.067 & -0.503 & 0.797 & [-0.164, 0.097] \\
Number of spiders: two+ (spider-present) & 0.033 & 0.067 & 0.503 & 0.797 & [-0.097, 0.164] \\
Color: color & -0.042 & 0.107 & -0.389 & 0.827 & [-0.251, 0.168] \\
Color: B/W & 0.042 & 0.107 & 0.389 & 0.827 & [-0.168, 0.251] \\
Size: middle (spider-present) & 0.019 & 0.052 & 0.373 & 0.827 & [-0.082, 0.121] \\
Eyes: non-visible (spider-present) & 0.014 & 0.046 & 0.304 & 0.833 & [-0.077, 0.105] \\
Eyes: visible (spider-present) & -0.014 & 0.046 & -0.304 & 0.833 & [-0.105, 0.077] \\
Size: small (spider-present) & 0.017 & 0.073 & 0.241 & 0.835 & [-0.125, 0.160] \\
Environment: human contact & -0.017 & 0.073 & -0.239 & 0.835 & [-0.160, 0.125] \\
Environment: nature & -0.011 & 0.058 & -0.195 & 0.845 & [-0.126, 0.103] \\
\end{longtable}

\FloatBarrier
\clearpage
\end{document}